\pdfoutput=1
\documentclass[11pt]{article}

\usepackage[]{EMNLP2022}

\usepackage{times}
\usepackage{latexsym}
\usepackage{CJKutf8}
\usepackage{amsmath}
\usepackage{verbatim}
\usepackage{booktabs}
\usepackage{bm}
\usepackage{amssymb}% http://ctan.org/pkg/amssymb
\usepackage{pifont}% http://ctan.org/pkg/pifont
\usepackage{stfloats}
\usepackage{multicol}
\usepackage{float}
\usepackage{graphicx}
\usepackage{subfigure}
\usepackage{multirow}
\usepackage{array}
\newcommand{\PreserveBackslash}[1]{\let\temp=\\#1\let\\=\temp}
\newcolumntype{C}[1]{>{\PreserveBackslash\centering}p{#1}}
\newcolumntype{R}[1]{>{\PreserveBackslash\raggedleft}p{#1}}
\newcolumntype{L}[1]{>{\PreserveBackslash\raggedright}p{#1}}
\usepackage{diagbox}
\usepackage{arydshln}
\usepackage{booktabs}
\usepackage[ruled,linesnumbered,lined,commentsnumbered]{algorithm2e}
\usepackage{mathtools}
\usepackage{amsfonts}
\usepackage{bbm}

\SetCommentSty{mycommfont}

\makeatletter
\def\hlinew#1{%
  \noalign{\ifnum0=`}\fi\hrule \@height #1 \futurelet
   \reserved@a\@xhline}
\makeatother

\usepackage[T1]{fontenc}
\usepackage[utf8]{inputenc}

\usepackage{microtype}

\title{Information-Transport-based Policy for Simultaneous Translation}

\author{Shaolei Zhang \textsuperscript{\rm 1,2},
    Yang Feng \textsuperscript{\rm 1,2}\thanks{ \ \ Corresponding author: Yang Feng.} \\
        \textsuperscript{\rm 1}{Key Laboratory of Intelligent Information Processing} \\ Institute of Computing Technology, Chinese Academy of Sciences (ICT/CAS) \\
    { \textsuperscript{\rm 2} {University of Chinese Academy of Sciences, Beijing, China}} \\
     \texttt{\{\href{mailto:zhangshaolei20z@ict.ac.cn}{zhangshaolei20z}, \href{mailto:fengyang@ict.ac.cn}{fengyang}\}@ict.ac.cn}  }

\begin{document}
\maketitle
\begin{abstract}
Simultaneous translation (ST) outputs translation while receiving the source inputs, and hence requires a policy to determine whether to translate a target token or wait for the next source token. The major challenge of ST is that each target token can only be translated based on the current received source tokens, where the received source information will directly affect the translation quality. So naturally, how much source information is received for the translation of the current target token is supposed to be the pivotal evidence for the ST policy to decide between translating and waiting. In this paper, we treat the translation as information transport from source to target and accordingly propose an \emph{Information-Transport-based Simultaneous Translation} (\emph{ITST}). ITST quantifies the transported information weight from each source token to the current target token, and then decides whether to translate the target token according to its accumulated received information. Experiments on both text-to-text ST and speech-to-text ST (a.k.a., streaming speech translation) tasks show that ITST outperforms strong baselines and achieves state-of-the-art performance\footnote{Code is available at \url{https://github.com/ictnlp/ITST}}.

\end{abstract}

\section{Introduction}

Simultaneous translation (ST) \cite{Cho2016,gu-etal-2017-learning,ma-etal-2019-stacl,Arivazhagan2019}, which outputs translation while receiving the streaming inputs, is essential for many real-time scenarios, such as simultaneous interpretation, online subtitles and live broadcasting. Compared with the conventional full-sentence machine translation (MT) \cite{NIPS2017_7181}, ST additionally requires a read/write policy to decide whether to wait for the next source input (a.k.a., READ) or generate a target token (a.k.a., WRITE).

The goal of ST is to achieve high-quality translation under low latency, however, the major challenge is that the low-latency requirement restricts the ST model to translating each target token only based on current received source tokens \cite{ma-etal-2019-stacl}. To mitigate the impact of this restriction on translation quality, ST needs a reasonable read/write policy to ensure that before translating, the received source information is sufficient to generate the current target token \cite{Arivazhagan2019}. To achieve this, read/write policy should measure the amount of received source information, if the received source information is sufficient for translation, the model translates a target token, otherwise the model waits for the next input.

\begin{figure}[t]
\centering
\includegraphics[width=2.95in]{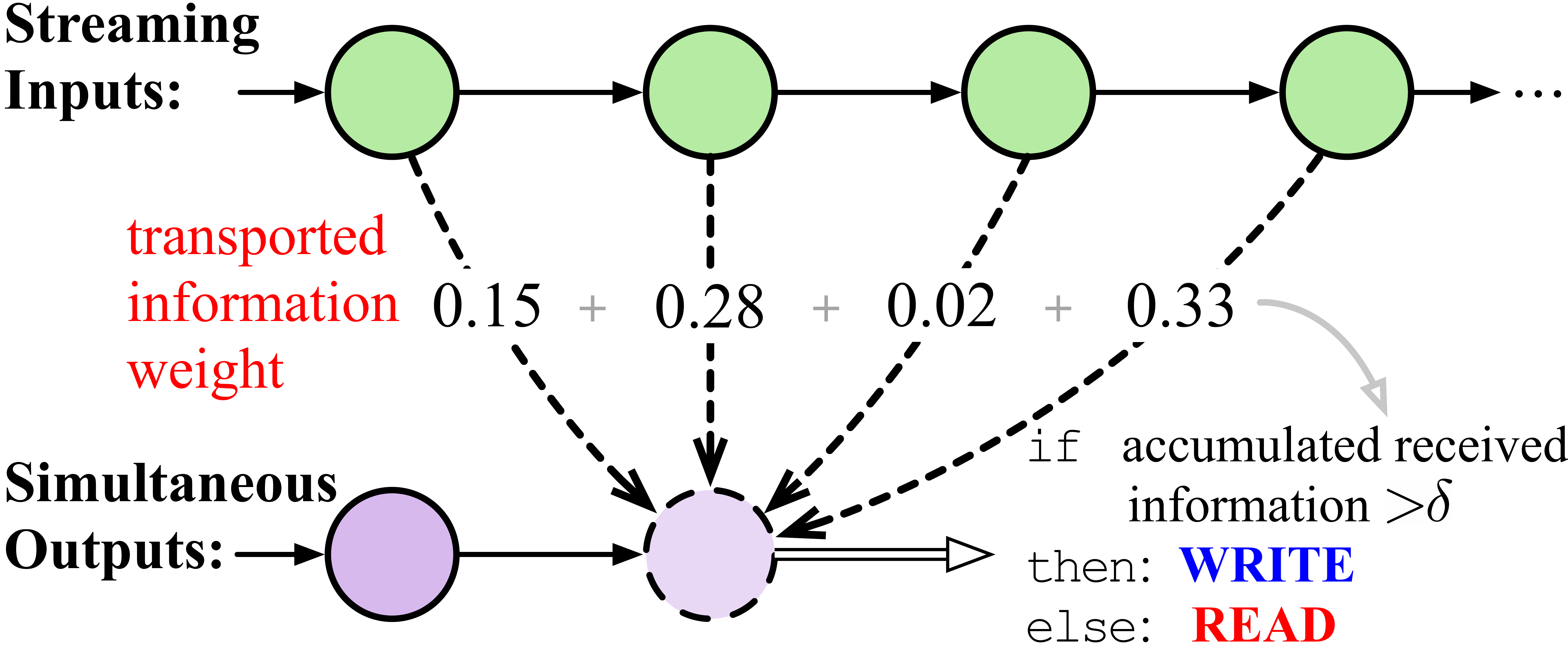}
\caption{Schematic diagram of ITST (e.g., $\delta=0.7$).}
\label{ill}
\end{figure}

However, previous read/write policies, involving fixed and adaptive, often lack an explicit measure of how much source information is received for the translation. Fixed policy decides READ/WRITE according to predefined rules \cite{ma-etal-2019-stacl,zhang-feng-2021-universal} and sometimes forces the model to start translating even though the received source information is insufficient, thereby affecting the translation quality. Adaptive policy can dynamically adjust READ/WRITE \cite{Arivazhagan2019,Ma2019a} to achieve better performance. However, previous adaptive policies often directly predict a variable based on the inputs to indicate READ/WRITE decision \cite{Arivazhagan2019,Ma2019a,miao-etal-2021-generative}, without explicitly modeling the amount of information that the received source tokens provide to the currently generated target token.

Under these grounds, we aim to develop a reasonable read/write policy that takes the received source information as evidence for READ/WRITE. For the ST process, source tokens provide information, while target tokens receive information and then perform translating, thereby the translation process can be treated as \emph{information transport} from source to target. Along this line, if we are well aware of how much information is transported from each source token to the target token, then it is natural to grasp the total information provided by the received source tokens for the current target token, thereby ensuring that the source information is sufficient for translation.

To this end, we propose \emph{Information-Transport-based Simultaneous Translation} (\emph{ITST}). Borrowing the idea from the optimal transport problem \cite{Villani2008OptimalTO}, ITST explicitly quantifies the transported information weight from each source token to the current target token during translation. Then, ITST starts translating after judging that the amount of information provided by received source tokens for the current target token has reached a sufficient proportion. As shown in the schematic diagram in Figure \ref{ill}, assuming that 70\% source information is sufficient for translation, ITST first quantifies the transport information weight from each source token to the current target token (e.g., $0.15,0.28,\cdots$). With the first three source tokens, the accumulated received information is 45\%, less than 70\%, then ITST selects READ. 
After receiving the fourth source token, the accumulated information received by the current target token becomes 78\%, thus ITST selects WRITE to translate the current target token. Experiments on both text-to-text and speech-to-text simultaneous translation tasks show that ITST outperforms strong baselines and achieves state-of-the-art performance.

\section{Background}
\label{sec:ST}

\textbf{Simultaneous Translation}\quad For the ST task, we denote the source sequence as $\mathbf{x}\!=\!\left ( x_{1},\cdots ,x_{J} \right )$ and the corresponding source hidden states as $\mathbf{z}\!=\!\left ( z_{1},\cdots ,z_{J} \right )$ with source length $J$. The model generates a target sequence $\mathbf{y}\!=\!\left ( y_{1},\cdots ,y_{I} \right )$ and the corresponding target hidden states $\mathbf{s}\!=\!\left ( s_{1},\cdots ,s_{I} \right )$ with target length $I$. Since ST model outputs translation while receiving the source inputs, we denote the number of received source tokens when translating $y_{i}$ as $g_{i}$. Then, the probability of generating $y_{i}$ is $p\left ( y_{i}\mid \mathbf{x}_{\leq g_{i}},\mathbf{y}_{< i} ; \bm{\theta }\right )$, where $\bm{\theta }$ is model parameters, $\mathbf{x}_{\leq g_{i}}$ is the first $g_{i}$ source tokens and $\mathbf{y}_{< i}$ is the previous target tokens. Accordingly, ST model is trained by minimizing the cross-entropy loss:
\begin{gather}
    \mathcal{L}_{ce}=-\sum_{i=1}^{I}\mathrm{log}\;p\left ( y^{\star}_{i}\mid \mathbf{x}_{\leq g_{i}},\mathbf{y}_{< i}; \bm{\theta } \right ),
\end{gather}
where $y^{\star}_{i}$ is the ground-truth target token.

\textbf{Cross-attention} Translation models often use cross-attention to measure the similarity of the target token and the source token \cite{NIPS2017_7181}, thereby weighting the source information \cite{wiegreffe-pinter-2019-attention}. Given the target hidden states $\mathbf{s}$ and source hidden states $\mathbf{z}$, the attention weight $\alpha _{ij}$ between $y_{i}$ and $x_{j}$ is calculated as:
\begin{gather}
\alpha_{ij} = \mathrm{softmax}\left(\frac{ s_{i} W^{Q}\left ( z_{j} W^{K} \right )^{\top}}{\sqrt{d_{k}}}\right),
\end{gather}
where $W^{Q}$ and $W^{K}$ are projection parameters, and $d_{k}$ is the dimension of inputs. Then the context vector $o_{i}$ is calculated as $o_{i}=\sum_{j=1}^{J} \alpha _{ij}\left ( z_{j}  W^{V}\right )$, where $W^{V}$ are projection parameters.

\section{The Proposed Method}
\label{sec:method}

We propose information-transport-based simultaneous translation (ITST) to explicitly measure the source information projected to the current generated target token. During the ST process, ITST models the information transport to grasp how much information is transported from each source token to the current target token (Sec.\ref{sec:Learning of Information Transport}). Then, ITST starts translating a target token after its accumulated received information is sufficient (Sec.\ref{sec:Policy}). Details of ITST are as follows.

\subsection{Information Transport}
\label{sec:Learning of Information Transport}

\textbf{Definition of Information Transport}\quad Borrowing the idea of optimal transport problem (OT) \cite{10.2307/1905523}, which aims to look for a transport matrix transforming a probability distribution into another while minimizing the cost of transport, we treat the translation process in ST as an information transport from source to target. We denote the information transport as the matrix $\mathbf{T}=\left ( T_{ij} \right )_{I\times J}$, where $T_{ij} \!\in\! (0,1)$ is the transported information weight from $x_{j}$ to $y_{i}$. Then, we assume that the total information received by each target token\footnote{Since the participation degree of each source token in translation is often different, we relax the constraints on total information provided by source token \cite{pmlr-v37-kusnerb15}.} for translation is $1$, i.e., $\sum_{j=1}^{J}T_{ij}=1$.

Under this definition, ITST quantifies the transported information weight $T_{ij}$ base on the current target hidden state $s_{i}$ and source hidden state $z_{j}$: 
\begin{gather}
    T_{ij}=\textrm{sigmoid}\left ( \frac{s_{i}V^{Q}\left (z_{j}V^{K}  \right )^{\top }}{\sqrt{d_{k}}} \right ) \label{eq5}
\end{gather}
where $V^{Q}$ and $V^{K}$ are learnable parameters.

\begin{figure*}[t]
\centering
\includegraphics[width=6.3in]{matrix2.pdf}
\caption{Schematic diagram of learning information transport matrix $\mathbf{T}$ from both translation and latency.}
\label{matrix}
\end{figure*}

\textbf{Constraints on Information Transport}\quad Similar to the OT problem, modeling information transport in translation also requires the transport costs to constrain the transported weights. Especially for ST, we should constrain information transport $\mathbf{T}$ from the aspects of translation and latency, where the translation constraints ensure that information transport can correctly reflect the translation process from source to target and the latency constraints regularize the information transport to avoid anomalous translation latency.

For \emph{translation constraints}, the information transport $\mathbf{T}$ should learn which source token contributes more to the translation of the current target token, i.e., reflecting the translation process. Fortunately, the cross-attention $\alpha _{ij}$ in the translation model is used to control the weight that source token $x_{j}$ provides to the target token $y_{i}$ \cite{abnar-zuidema-2020-quantifying,chen-etal-2020-accurate,zhang-feng-2021-modeling-concentrated}, so we integrate information transport into the cross-attention. As shown in Figure \ref{matrix}(a), we multiply $T_{ij}$ with cross-attention $\alpha _{ij}$ and then normalize to get final attention $\beta _{ij}$:
\begin{gather}
    \hat{\beta }_{ij}=\alpha _{ij}\times T_{ij}\;,\;\;\;\;\beta _{ij}=\hat{\beta }_{ij}/\sum_{j=1}^{J}\hat{\beta }_{ij} .
\end{gather}
Then the context vector is calculated as $o_{i}\!=\!\sum_{j=1}^{J}\beta _{ij}\left ( z_{j}W^{V} \right )$. In this way, the information transport $\mathbf{T}$ can be jointly learned with the cross-attention in the translation process through the original cross-entropy loss $\mathcal{L}_{ce}$.

For \emph{latency constraints}, the information transport $\mathbf{T}$ will affect the translation latency, since the model should start translating after receiving a certain amount of information. Specifically, for the current target token, if too much information is provided by the source tokens lagging behind, waiting for those source tokens will cause high latency. While too much information provided by the front source tokens will make the model prematurely start translating, resulting in extremely low latency and poor translation quality \cite{laf}. Therefore, we aim to avoid too much information weight being transported from source tokens that are located too early or too late compared to the position of the current target token, thereby getting a suitable latency.

To this end, we introduce a latency cost matrix $\mathbf{C}\!=\!\left ( C_{ij} \right )_{I\times J}$ in the diagonal form to softly regularize the information transport, where $C_{ij}$ is the latency cost of transporting information from $x_{j}$ to $y_{i}$, related to their relative offset:
\begin{gather}
    C_{ij}=\frac{1}{I\!\times \!J}\left (\! \textrm{max}\left (  \left|j-i\times \frac{J}{I} \right|-\xi  ,0 \right )\!\right ). \label{eq9}
\end{gather}
$\left|j-i\times \frac{J}{I} \right|$ is the relative offset between $x_{j}$ and $y_{i}$. $\xi$ is a hyperparameter to control the acceptable offset (i.e. inside transports cost 0), and we set $\xi\! =\!1$ in our experiments. As an example of the latency cost matrix shown in Figure \ref{matrix}(c), the transported weights cost 0 when the relative offset less than 1, and the cost of other transports is positively related to the offset. We will compare different settings of the latency cost in Sec.\ref{sec:ab} and Appendix \ref{app:latency cost}. 

Given the latency cost matrix $\mathbf{C}$, the latency loss $\mathcal{L}_{latency}$ of information transport $\mathbf{T}$ is:
\begin{gather}
    \mathcal{L}_{latency}=\sum_{i=1}^{I}\sum_{j=1}^{J}\;T_{ij}\times C_{ij}. \label{eq7}
\end{gather}

\textbf{Learning Objective}\quad Accordingly, the learning of ST model $\bm{\theta}$ with the proposed information transport $\mathbf{T}$ can be formalized as:
\begin{align}
    \underset{\bm{\theta} ,\mathbf{T}}{\textrm{min}} \;\;\mathcal{L}_{ce}& + \mathcal{L}_{latency}   \\
    \textrm{s.t.}\;\;\;\;\sum_{j=1}^{J}T_{ij}=1,&\;\;\,\forall\;   1\!\leq \!i\!\leq \!I  \label{eq_10} \\
    T_{ij} \geq 0,&\;\;\,\forall\;   1\!\leq \!i\!\leq \!I, 1\!\leq\! j\!\leq \!J  \label{eq_11}
\end{align}
Eq.(\ref{eq_10}) constrains the total information transported to each target token to be 1 (refer to the definition), and Eq.(\ref{eq_11}) constrains the transported weights to be positive, realized by $\mathrm{sigmoid}\!\left ( \cdot  \right )$ in Eq.(\ref{eq5}). Then, we convert the normalization constraints of $T_{ij}$ in Eq.(\ref{eq_10}) into the following regular term:
\begin{gather}
    \mathcal{L}_{norm}=\sum_{i=1}^{I} \left\|\sum_{j=1}^{J}T_{ij}-1  \right\|_{2}. \label{eq11}
\end{gather}
Therefore, the total loss $\mathcal{L}_{ITST}$ is calculated as:
\begin{gather}
    \mathcal{L}_{ITST}=\mathcal{L}_{ce} + \mathcal{L}_{latency} +  \mathcal{L}_{norm}.
\end{gather}

\subsection{Information Transport based Policy}
\label{sec:Policy}

\begin{algorithm}[t]
\caption{Read/Write Policy of ITST}\label{algorithm}
  \SetKwData{Training}{Training}\SetKwData{This}{this}\SetKwData{Up}{up}
  \SetKwFunction{getInputTestLagging}{getInputTestLagging}\SetKwFunction{SampleFrom}{SampleFrom}
  \SetKwInOut{Input}{Input}\SetKwInOut{Output}{Output}
  \Input{Streaming inputs $\mathbf{x}$, Threshold $\delta$, $i=1$, $j=1$, $y_{0}=\left \langle \mathrm{BOS} \right \rangle$}
  \Output{Target outputs $\mathbf{y}$}
  \While{$y_{i-1}\neq\left \langle  \mathrm{EOS} \right \rangle$}{
    calculate \emph{information transport} $\mathbf{T}=(T_{i1},\cdots,T_{ij})$ as Eq.(\ref{eq5}); \\
    \eIf(\tcp*[f]{$\!\!\!\triangleright $WRITE}){$\;\;\sum_{l=1}^{j}\!T_{il}\geq\delta\;\;$}{
      translate $y_{i}$ with $\left(x_{1},\cdots,x_{j}\right)$; \\
      $i\leftarrow i+1$;
    }
    (\tcp*[f]{$\!\!\!\triangleright $READ}){
      wait for next source input $x_{j+1}$; \\
      $j\leftarrow j+1$; \\
    }
    
    }
\end{algorithm}

\textbf{Read/Write Policy}\quad After grasping the information transported from each source token to the current target token, we propose an information-transport-based policy accordingly. With streaming inputs, ITST receives source tokens one by one and transports their information to the current target token, and then ITST starts translating when the accumulated received information is sufficient. To obtain a controllable latency \cite{ma-etal-2019-stacl} during testing, a threshold $\delta$ is introduced to indicate how much proportion of source information is sufficient for translation. Therefore, as shown in Algorithm \ref{algorithm}, ITST selects WRITE after the accumulated received source information of the current target token $\sum_{l=1}^{j}\!T_{il}$ is greater than the threshold $\delta$, otherwise ITST selects READ. 

ITST can perform translating under different latency by adjusting the threshold $\delta$. With larger $\delta$, ITST tends to wait for more transported information, so the latency becomes higher; otherwise, the latency becomes lower with smaller $\delta$.

\textbf{Curriculum-based Training}\quad Besides a reasonable read/write policy, ST model also requires the capability of translating based on incomplete source information. Therefore, we apply the threshold in training as well, denoted as $\delta_{train}$, and accordingly mask out the rest of source tokens when accumulated information of each target token exceeds $\delta_{train}$. Formally, given $\delta_{train}$, $y_{i}$ is translated based on the first $g_{i}$ source tokens, where $g_{i}$ is:
\begin{gather}
    g_{i}=\underset{j}{\textrm{argmin}}\sum_{l=1}^{j}T_{il}\geq \delta_{train}.
\end{gather}
Then, we mask out the source token $x_{j}$ that $j>g_{i}$ during training to simulate the streaming inputs.

Regarding how to set $\delta_{train}$ during training, unlike previous methods that train multiple separate ST models for different thresholds \cite{ma-etal-2019-stacl,Ma2019a} or randomly sample different thresholds \cite{multipath,zhang-feng-2021-universal}, we propose \emph{curriculum-based training} for ITST to train one universal model that can perform ST under arbitrary latency (various $\delta$ during testing).

The proposed curriculum-based training follows an easy-to-hard schedule. At the beginning of training, we let the model preferentially focus on the learning of translation and information transport under the richer source information, Then, we gradually reduce the source information as the training progresses to let the ST model learn to translate with incomplete source inputs. Therefore, $\delta_{train}$ is dynamically adjusted according to an exponential-decaying schedule during training:
\begin{gather}
    \delta_{train} \!=\!\delta_{min}\!+\!\left (1\!\!-\!\delta_{min}\right )\!\times\! \mathrm{exp}\!\left (\! -\frac{N_{update}}{d} \!\right ), \label{eq17}
\end{gather}
where $N_{update}$ is update steps, and $d$ is a hyperparameter to control the decaying degree. $\delta_{min}$ is the minimum amount of information required, and we set $\delta_{min}\!=\!0.5$ in the experiments. Thus, during training, the information received by each target token gradually decays from 100\% to 50\%.

\section{Experiments}
\subsection{Datasets}
We conduct experiments on both text-to-text ST and speech-to-text ST tasks.

$\bullet$\quad Text-to-text ST (T2T-ST)

\textbf{IWSLT15\footnote{\url{nlp.stanford.edu/projects/nmt/}} English $\!\rightarrow \!$ Vietnamese (En$\rightarrow$Vi)} (133K pairs) \cite{iwslt2015}\quad We use TED tst2012 as the validation set (1553 pairs) and TED tst2013 as the test set (1268 pairs). Following the previous setting \cite{LinearTime,Ma2019a}, we replace tokens that the frequency less than 5 by $\left \langle unk \right \rangle$, and the vocabulary sizes are 17K and 7.7K for English and Vietnamese respectively.

\textbf{WMT15\footnote{\url{www.statmt.org/wmt15/}} German $\!\rightarrow\! $ English (De$\rightarrow$En)} (4.5M pairs)\quad We use newstest2013 as the validation set (3000 pairs) and newstest2015 as the test set (2169 pairs). 32K BPE \cite{sennrich-etal-2016-neural} is applied and the vocabulary is shared across languages.

$\bullet$\quad Speech-to-text ST (S2T-ST)

\textbf{MuST-C\footnote{\url{https://ict.fbk.eu/must-c}} } \textbf{English $\!\rightarrow\! $ German (En$\rightarrow$De)} (234K pairs) and \textbf{English $\!\rightarrow\! $ Spanish (En$\rightarrow$Es)} (270K pairs) \cite{di-gangi-etal-2019-must}.\quad We use \texttt{dev} as validation set (1423 pairs for En$\rightarrow$De, 1316 pairs for En$\rightarrow$Es) and use \texttt{tst-COMMON} as test set (2641 pairs for En$\rightarrow$De, 2502 pairs for En$\rightarrow$Es), respectively. Following \citet{ma-etal-2020-simulmt}, we use Kaldi \cite{Povey:192584} to extract 80-dimensional log-mel filter bank features for speech, computed with a 25$ms$ window size and a 10$ms$ window shift, and we use SentencePiece \cite{kudo-richardson-2018-sentencepiece} to generate a unigram vocabulary of size $10000$ respectively for source and target text.

\subsection{Experimental Settings}
We conduct experiments on the following systems. All implementations are based on Transformer \cite{NIPS2017_7181} and adapted from Fairseq Library \cite{ott-etal-2019-fairseq}.

{\bf Offline}\quad Full-sentence MT \cite{NIPS2017_7181}, which waits for the complete source inputs and then starts translating.

{\bf Wait-k}\quad Wait-k policy \cite{ma-etal-2019-stacl}, the most widely used fixed policy, which first READ $k$ source tokens, and then alternately READ one token and WRITE one token.

{\bf Multipath Wait-k}\quad An efficient training for wait-k \cite{multipath}, which randomly samples different $k$ between batches during training.

{\bf Adaptive Wait-k}\quad A heuristic composition of multiple wait-k models ($k\!=\!1\cdots13$) \cite{zheng-etal-2020-simultaneous}, which decides whether to translate according to the generating probabilities of wait-k models.

{\bf{MoE Wait-k}\footnote{\url{github.com/ictnlp/MoE-Waitk}}}\quad Mixture-of-experts wait-k policy \cite{zhang-feng-2021-universal}, the SOTA fixed policy, which applies multiple experts to learn multiple wait-k policies during training.

{\bf{MMA}\footnote{\url{github.com/pytorch/fairseq/tree/master/examples/simultaneous_translation}}}\quad Monotonic multi-head attention \cite{Ma2019a}, which predicts a Bernoulli variable to decide READ/WRITE, and the Bernoulli variable is jointly learning with multi-head attention.

{\bf GSiMT}\quad Generative ST \cite{miao-etal-2021-generative}, which also predicts a Bernoulli variable to decide READ/WRITE and the variable is trained with a generative framework via dynamic programming.

{\bf RealTranS}\quad  End-to-end simultaneous speech translation with Wait-K-Stride-N strategy \cite{zeng-etal-2021-realtrans}, which waits for $N$ frame at each step.

{\bf MoSST}\quad Monotonic-segmented streaming speech translation \cite{dong-etal-2022-learning}, which uses integrate-and-firing method to segment the speech.

{\bf ITST}\quad The proposed method in Sec.\ref{sec:method}.

\textbf{T2T-ST Settings}\quad We apply Transformer-Small (4 heads) for En$\rightarrow$Vi and Transformer-Base/Big (8/16 heads) for De$\rightarrow$En. Note that we apply the unidirectional encoder for Transformer to enable simultaneous decoding. Since GSiMT involves dynamic programming which makes its training expensive, we report GSiMT on WMT15 De$\rightarrow$En (Base) \cite{miao-etal-2021-generative}. For T2T-ST evaluation, we report BLEU \cite{papineni-etal-2002-bleu} for translation quality and Average Lagging (AL, token) \cite{ma-etal-2019-stacl} for latency. We also give the results with SacreBLEU in Appendix \ref{sec:Numerical Results}.

\begin{figure*}[t]
\centering
\subfigure[En$\rightarrow$Vi, Transformer-Small]{
\includegraphics[width=1.92in]{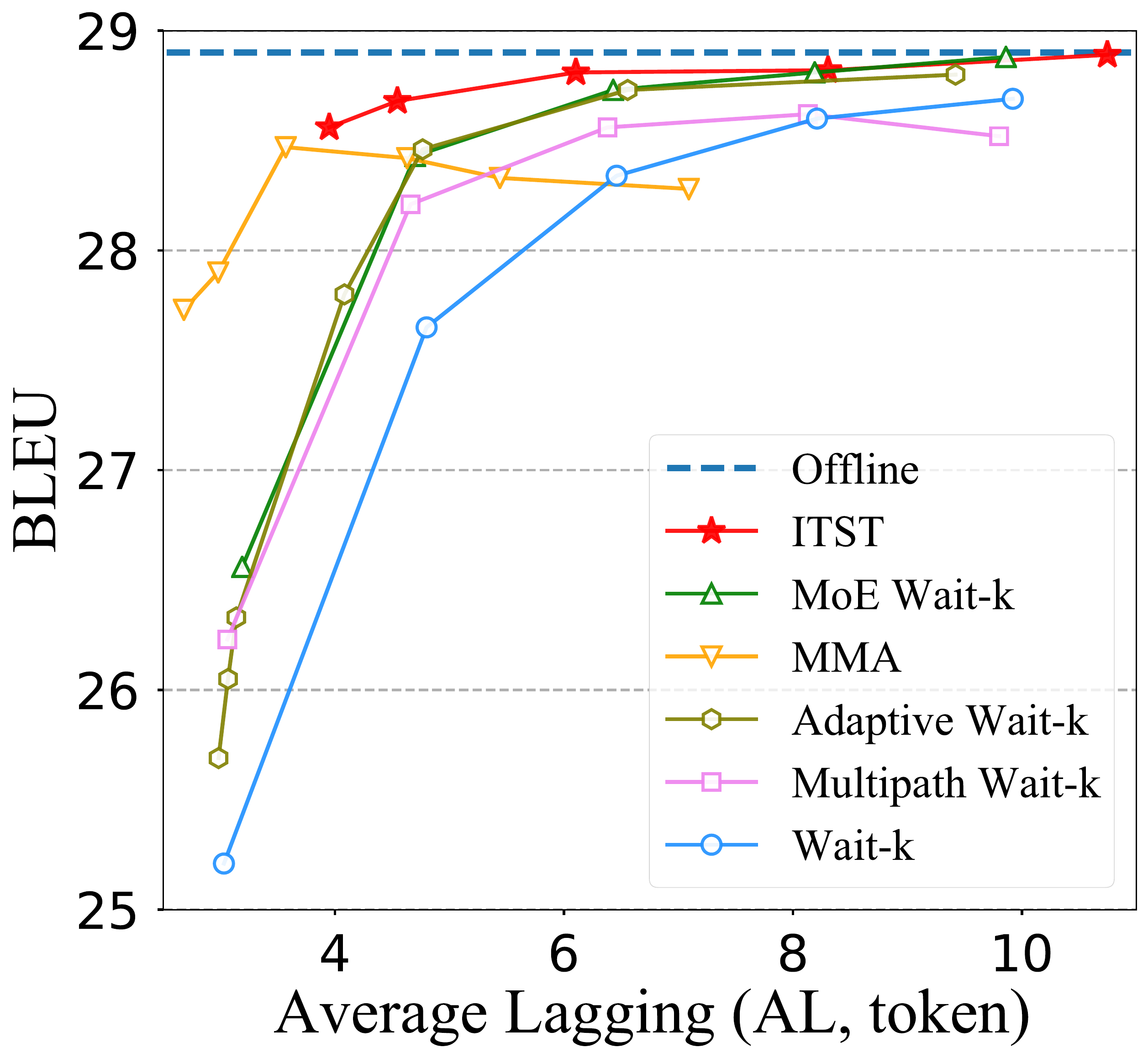}
}\hspace{2.6mm}
\subfigure[De$\rightarrow$En, Transformer-Base]{
\includegraphics[width=1.92in]{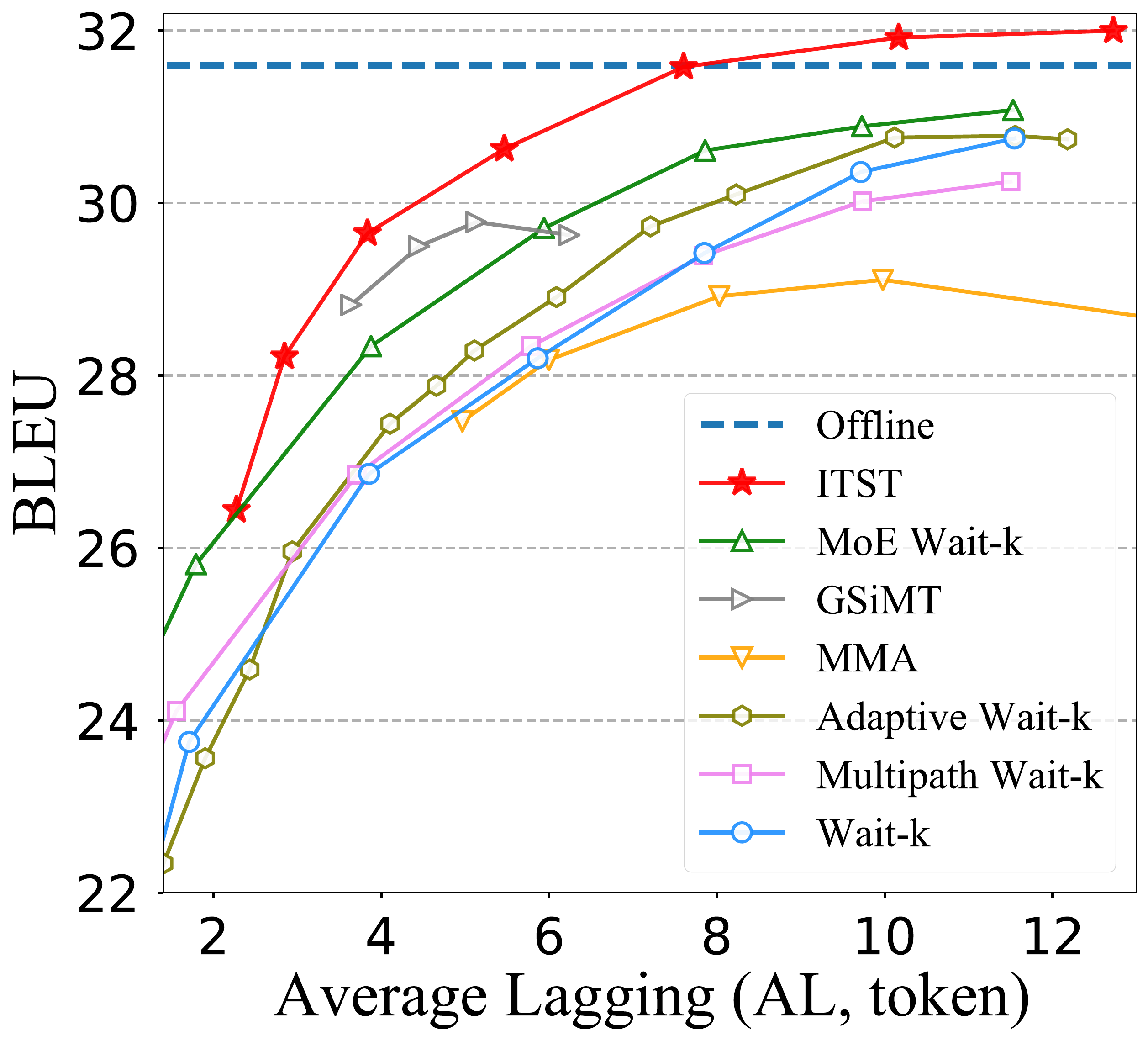}
}\hspace{2.6mm}
\subfigure[De$\rightarrow$En, Transformer-Big]{
\includegraphics[width=1.92in]{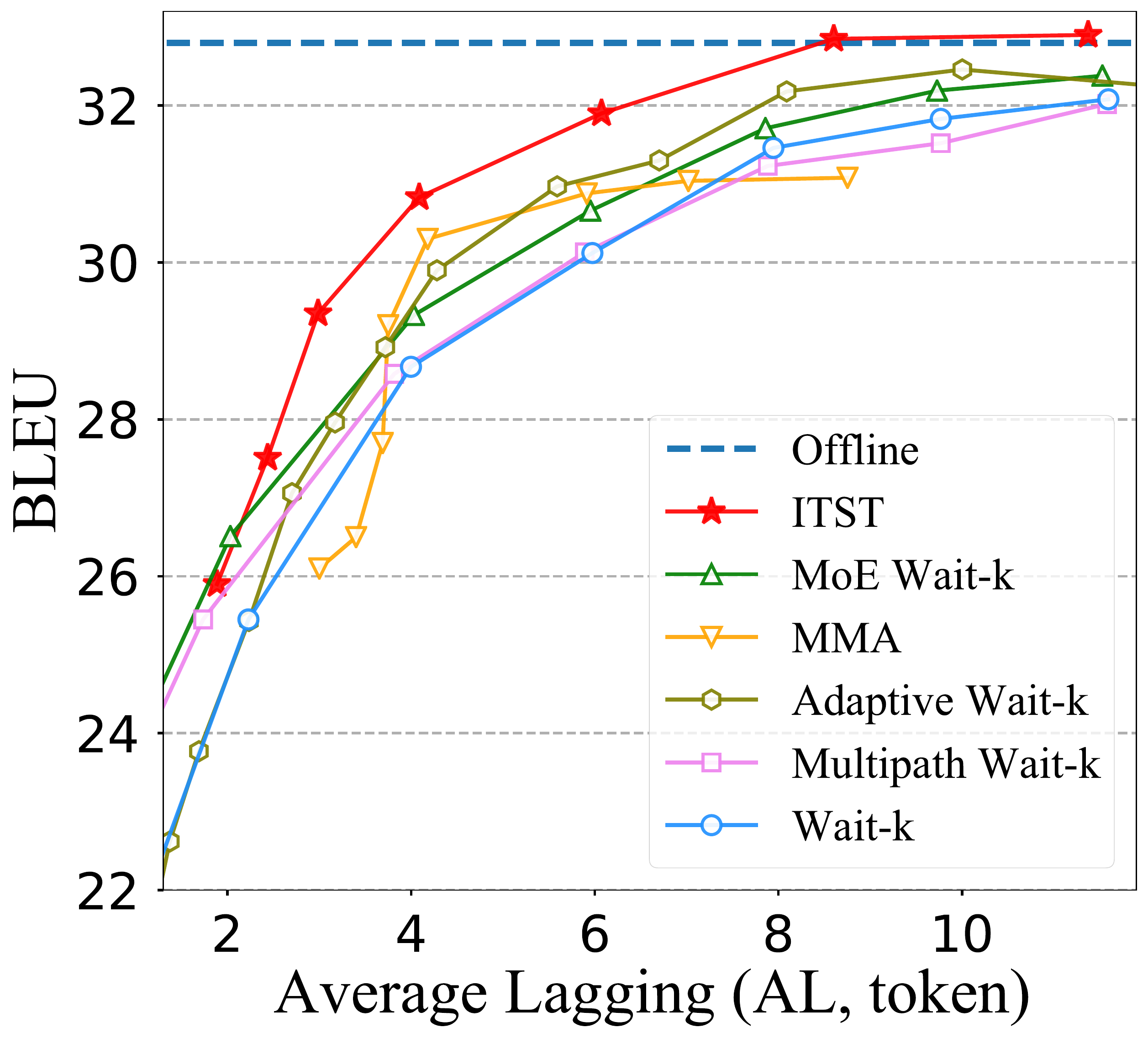}
}
\vspace{-4mm}
\caption{Text-to-text ST results of translation quality v.s. latency (token) on En$\rightarrow$Vi (Small), De$\rightarrow$En (Base, Big).}
\label{t2tmain}

\end{figure*}
\begin{figure*}[]
\begin{minipage}[t]{.64\linewidth}
\vspace{-2mm}
\subfigure[En$\rightarrow$De]{
\includegraphics[width=1.83in]{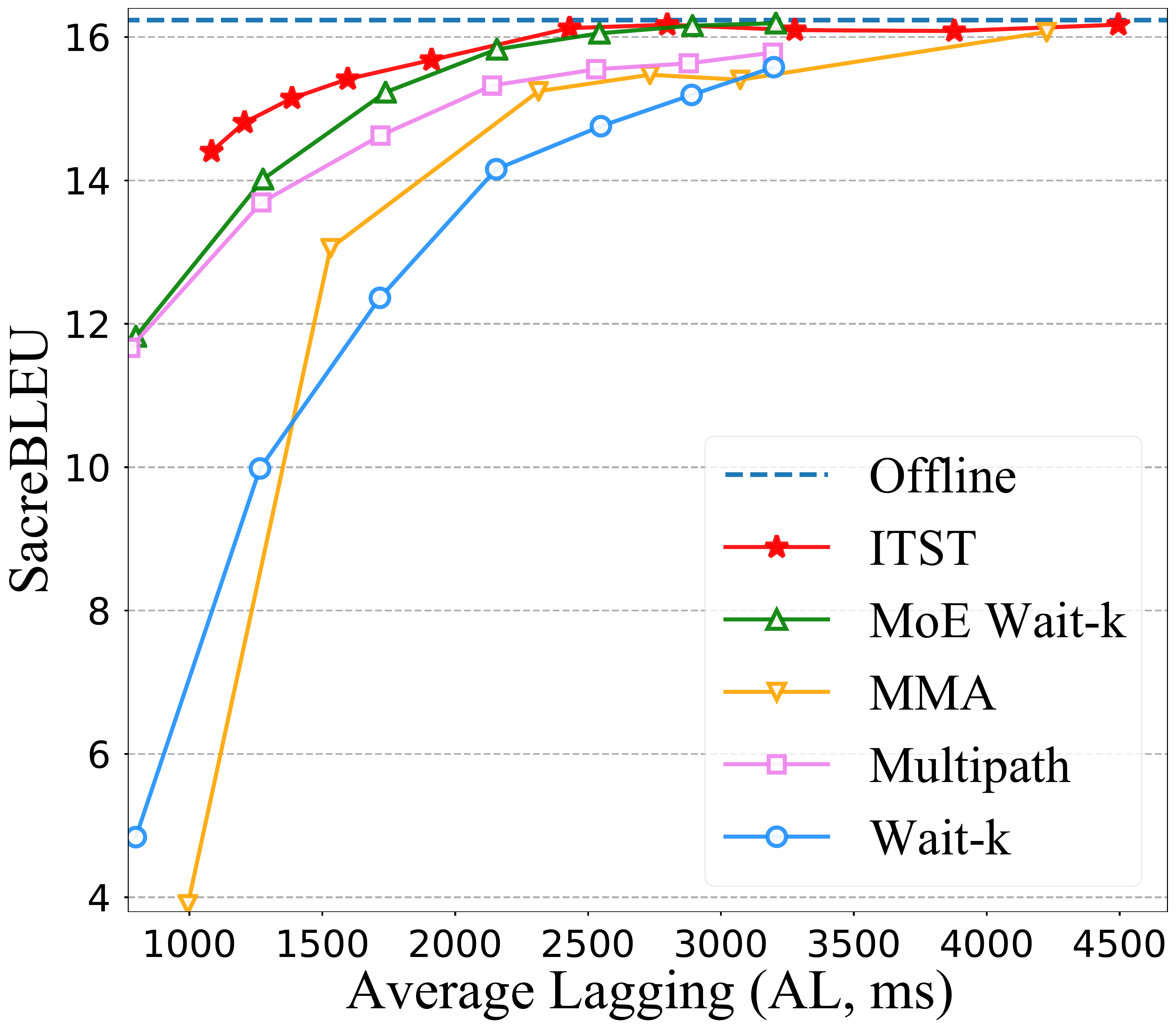}
}
\subfigure[En$\rightarrow$Es]{
\includegraphics[width=1.83in]{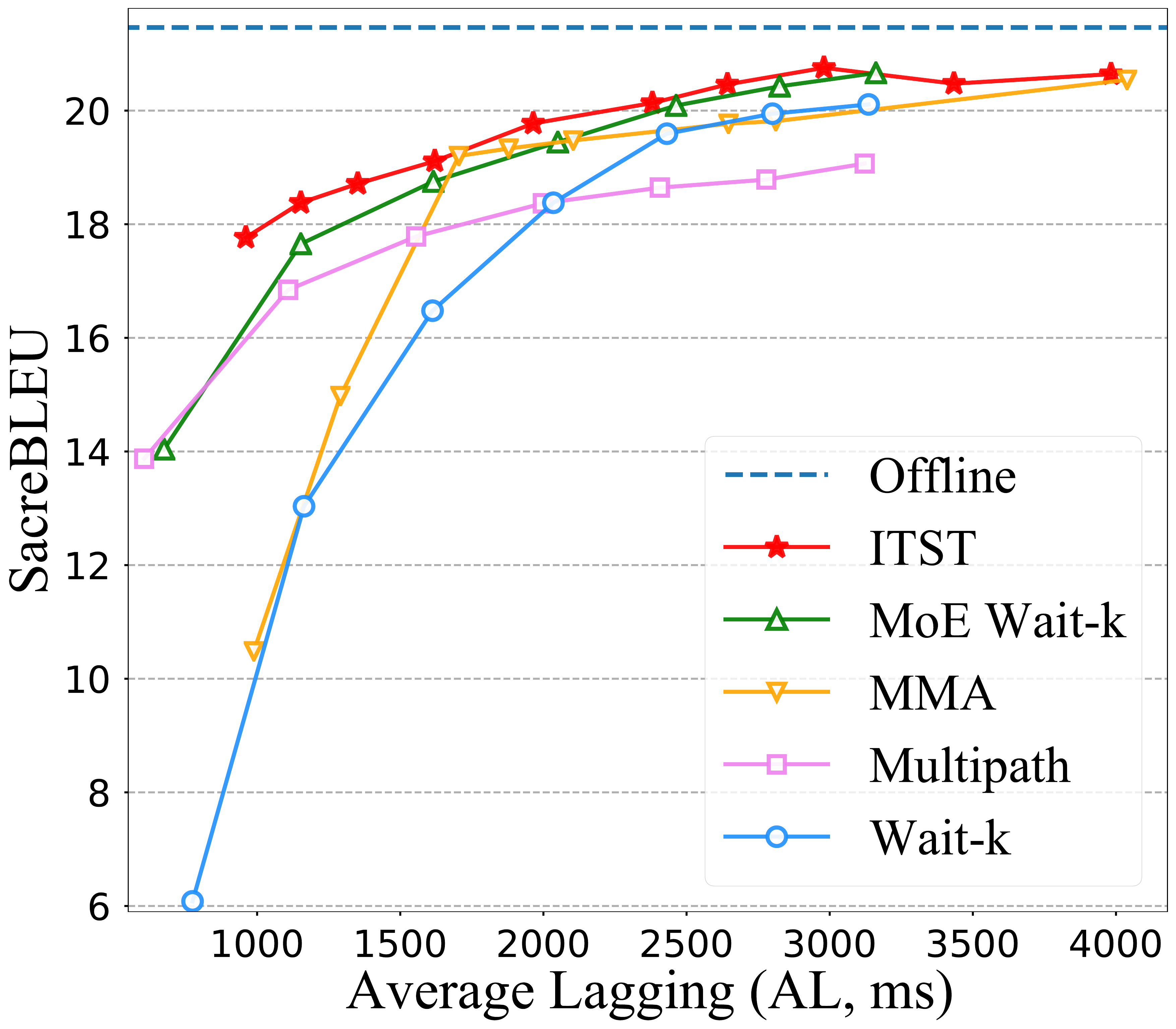}
}
\vspace{-4mm}
\caption{Speech-to-text ST results of translation quality v.s. latency ($ms$) on MuST-C En$\rightarrow$De and En$\rightarrow$Es with fixed pre-decision of 280$ms$. }
\label{s2tmain}
\end{minipage}
\hspace{5mm}
\begin{minipage}[t]{.31\linewidth}
\vspace{0mm}
\centering
\includegraphics[width=1.83in]{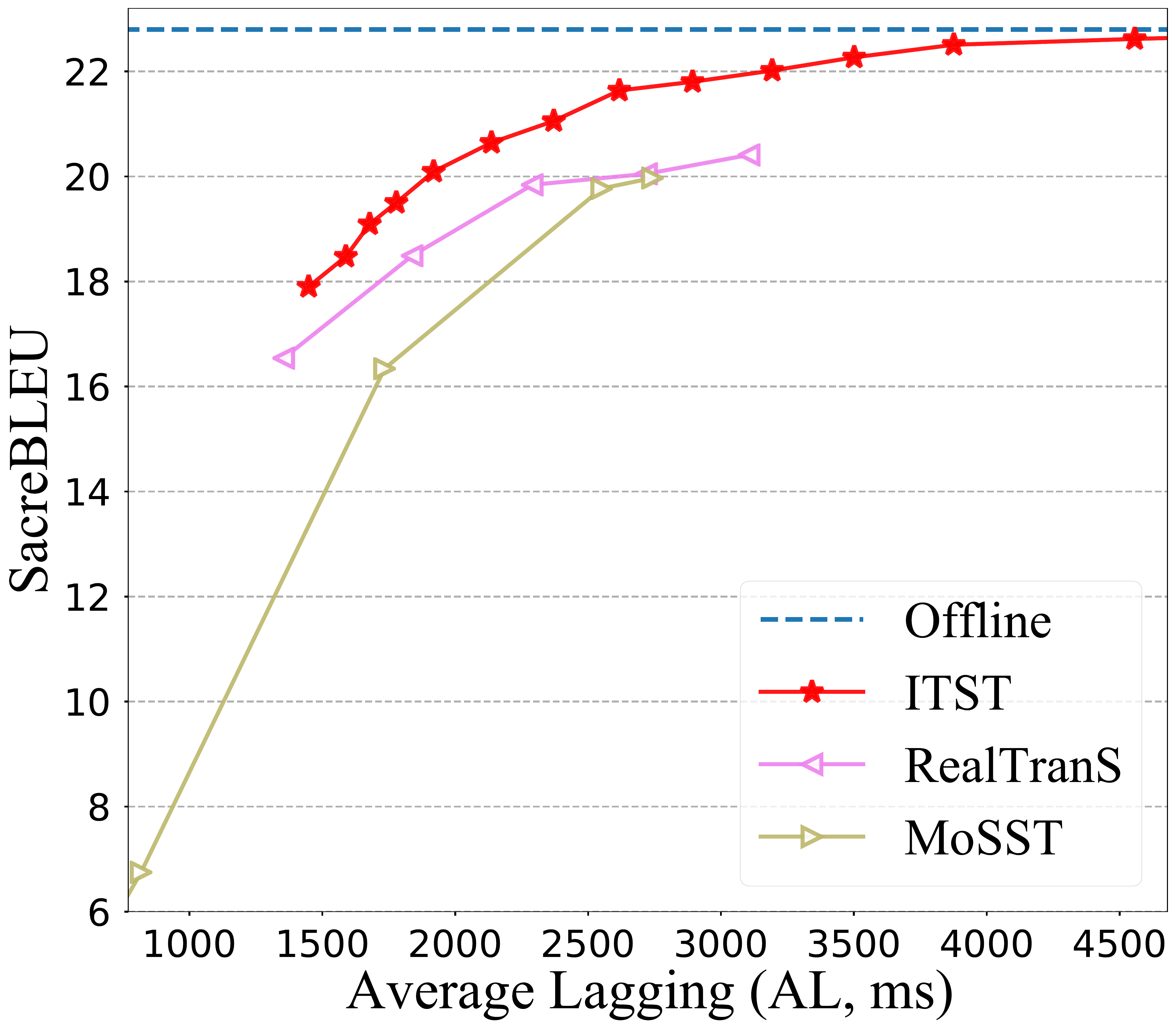}
\vspace{-1.4mm}
\caption{Speech-to-text ST results on MuST-C En$\rightarrow$De with flexible pre-decision on each speech frame.}
\label{s2tmain2}
\end{minipage}
\end{figure*}

\textbf{S2T-ST Settings}\quad The proposed ITST can perform end-to-end speech-to-text ST in two manners: \emph{fixed pre-decision} and \emph{flexible pre-decision} \cite{ma-etal-2020-simulmt}. For \emph{fixed pre-decision}, following \citet{ma-etal-2020-simulmt}, we apply ConvTransformer-Espnet (4 heads) \cite{inaguma-etal-2020-espnet} for both En$\rightarrow$De and En$\rightarrow$Es, which adds a 3-layer convolutional network before the encoder to capture the speech features. Note that the encoder is also unidirectional for simultaneous decoding. The convolutional layers and encoder are initialized from the pre-trained ASR task. All systems make a fixed pre-decision of READ/WRITE every 7 source tokens (i.e., every 280$ms$). 
For \emph{flexible pre-decision}, we use a pre-trained Wav2Vec2 module\footnote{\url{dl.fbaipublicfiles.com/fairseq/wav2vec/wav2vec_small.pt}} \cite{NEURIPS2020_92d1e1eb} to capture the speech features instead of using filter bank features, and a Transformer-Base follows to perform translating. To enable simultaneous decoding, we turn Wav2Vec2.0 into unidirectional type\footnote{\textbf{Unidirectional Wav2Vec2.0}: Turning the Transformer blocks in Wav2Vec2.0 into unidirectional (add the causal mask), and freeze the parameters of convolutional layers.} and apply unidirectional encoder for Transformer-Base. The model is trained by multi-task learning of ASR and ST tasks \cite{anastasopoulos-chiang-2018-tied,dong-etal-2022-learning}. When deciding READ/WRITE with the flexible pre-decision, ITST quantifies the transported information weight from each speech frame to the target token and then makes a decision. For S2T-ST evaluation, we apply SimulEval\footnote{\url{github.com/facebookresearch/SimulEval}} \cite{ma-etal-2020-simuleval} to report SacreBLEU \citep{post-2018-call} for translation quality and Average Lagging (AL, $ms$) for latency. 

All T2T and S2T systems apply greedy search.

\subsection{Main Results}

\textbf{Text-to-Text ST}\quad As shown in Figure \ref{t2tmain}, ITST outperforms previous methods under all latency and achieves state-of-the-art performance. Compared with fixed policies `Wait-k' and `MoE Wait-k', ITST dynamically adjusts READ/WRITE based on the amount of received source information instead of simply considering the token number, in which balancing at information level rather than token level is more reasonable for read/write policy \cite{wait-info} and thereby bring notable improvement. Compared with adaptive policies `MMA' and `GSiMT', ITST performs better and more stable. To decide READ/WRITE, previous adaptive policies directly predict decisions based on the last source token and target token \cite{Ma2019a,miao-etal-2021-generative}, while ITST explicitly measures the amount of accumulated received source information through the information transport and makes more reasonable decisions accordingly, thereby achieving better performance. Besides, previous adaptive policies previous adaptive policies often train multiple models for different latency, which sometimes leads to a decrease in translation quality under high latency \cite{Ma2019a,miao-etal-2021-generative}. The proposed curriculum-based training follows an easy-to-hard schedule and thus helps ST model achieve more stable performance under different latency. 

It is worth mentioning that ITST even outperforms the Offline model on De$\rightarrow$En (Base) when AL$>$8. This is because in addition to ITST providing a more reasonable read/write policy, modeling information transport (IT) in translation task can directly improve the translation quality. We will study the improvement on full-sentence MT brought by information transport in Sec.\ref{sec:full}.

\textbf{Speech-to-Text ST}\quad For fixed pre-decision, as shown in Figure \ref{s2tmain}, ITST outperforms previous methods on S2T-ST, especially improving about 10 BLEU compared with `Wait-k' and `MMA' under low latency (AL$<\!1000ms$). When the latency is about $2500ms$, ITST even achieves similar translation quality with full-sentence MT on En$\rightarrow$De. For flexible pre-decision, ITST outperforms the previous `RealTranS' and `MoSST', achieving the best trade-off between translation quality and latency. ITST models the information transport from each speech frame to the target token, thereby achieving a more reasonable and flexible read/write policy.

\section{Analysis}
We conduct extensive analyses to study the specific improvement of ITST. Unless otherwise specified, all results are reported on De$\rightarrow$En (Base) T2T-ST.

\subsection{Ablation Study}
\label{sec:ab}

\begin{figure}[t]
\centering
\subfigure[Effect of two constraints]{
\includegraphics[width=1.46in]{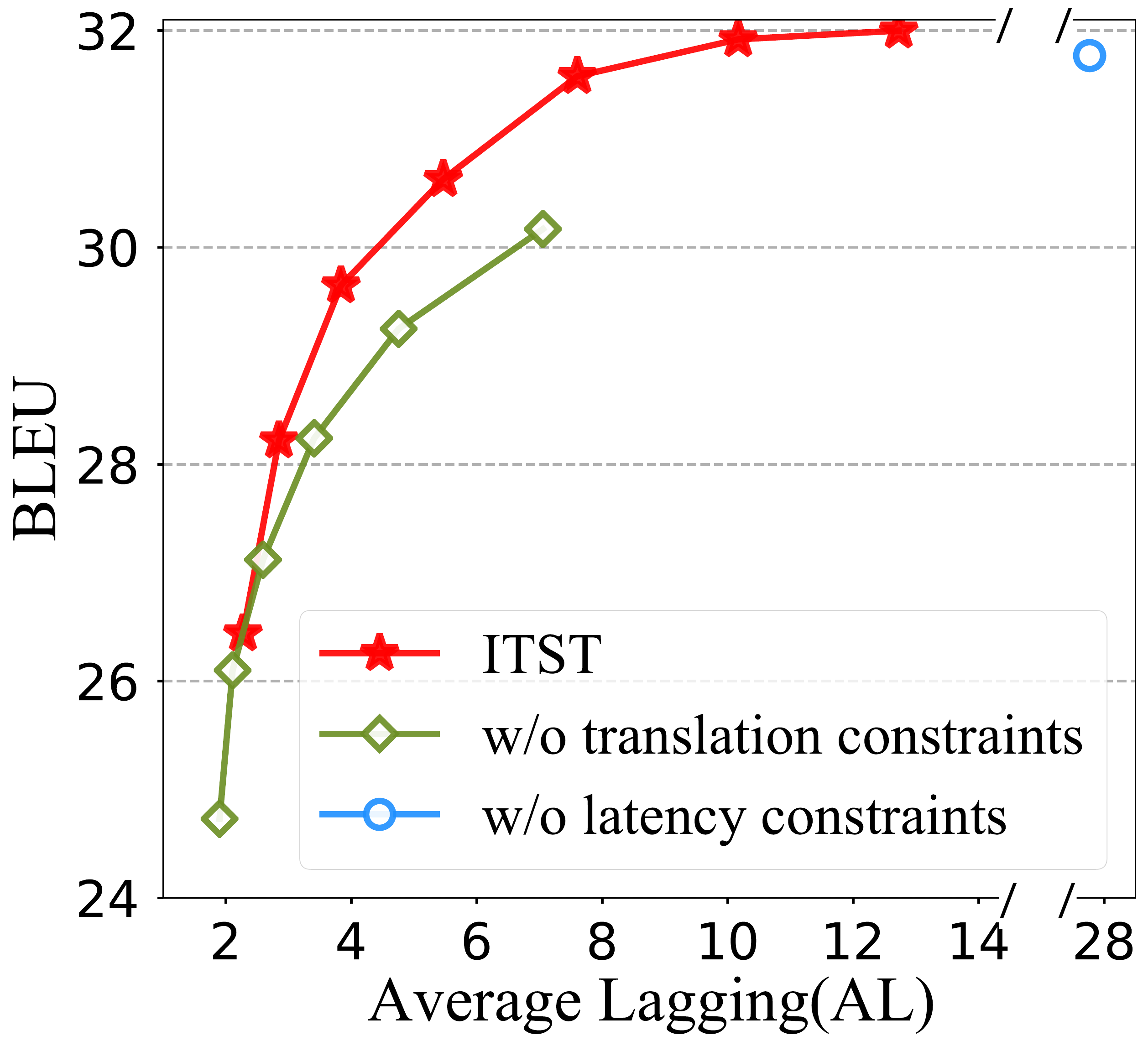} \label{ablation_a}
}\hspace{-2mm}
\subfigure[Various latency cost $C_{ij}$]{
\includegraphics[width=1.46in]{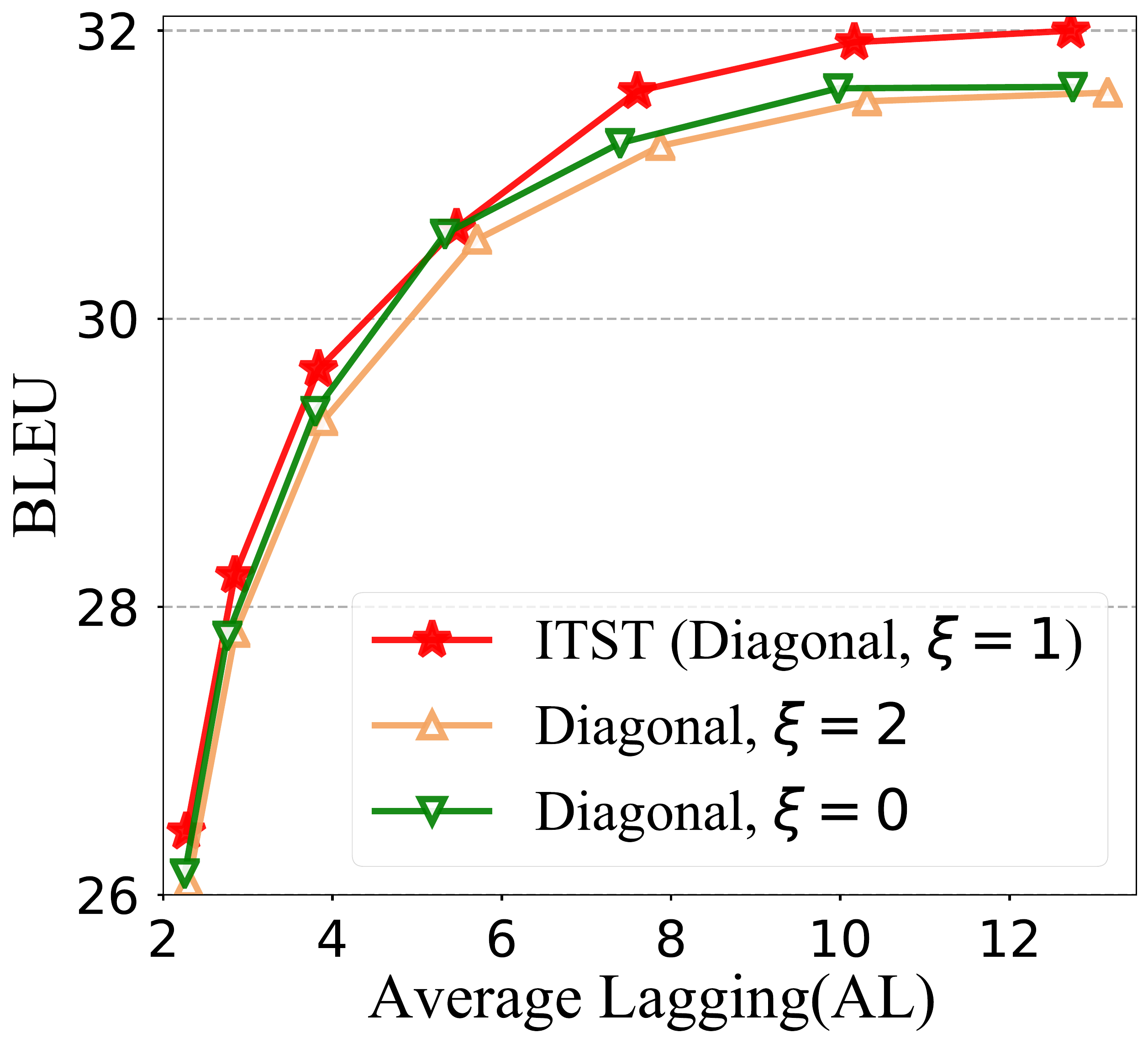} \label{ablation_b}
}
\subfigure[Normalization of information transport ($\sum_{j=1}^{J}T_{ij}$)]{
\includegraphics[width=2.93in]{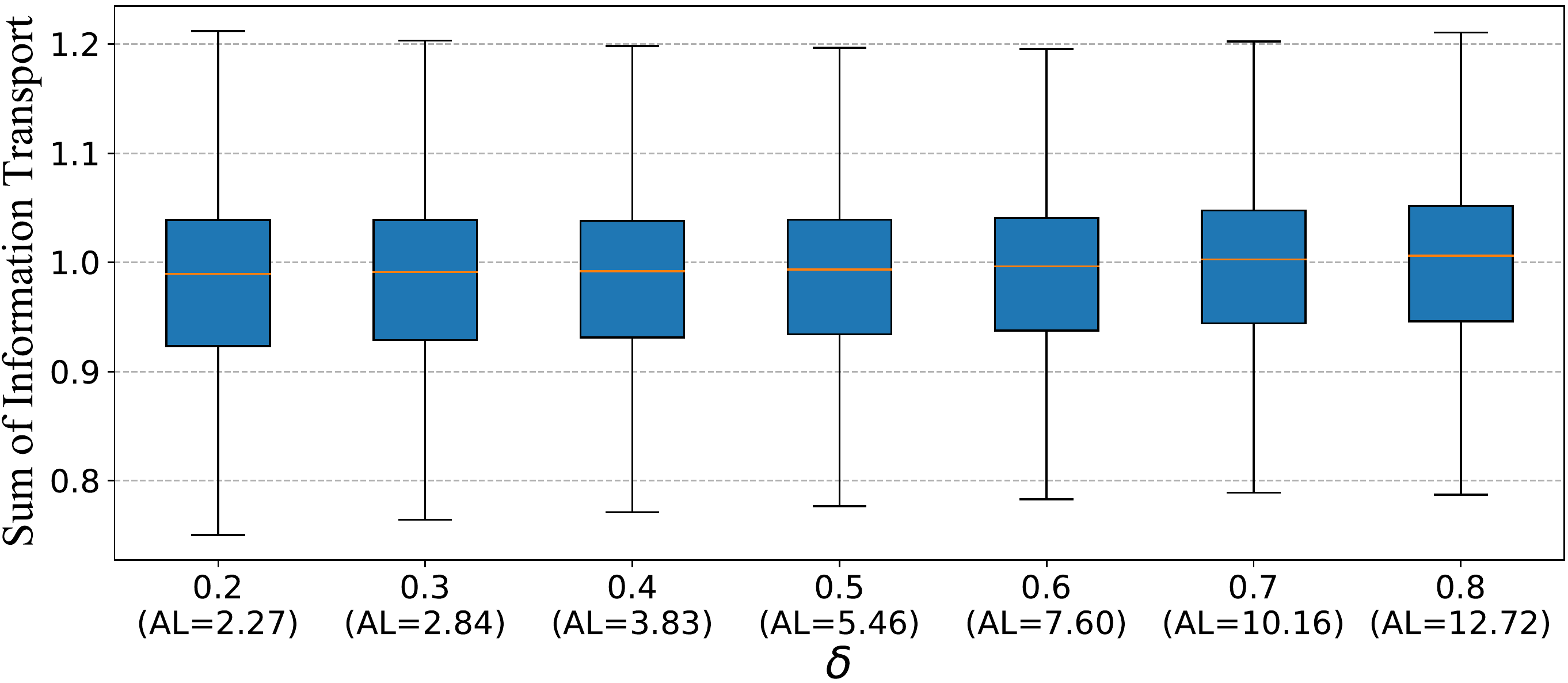} \label{ablation_c}
}
\vspace{-2.6mm}
\caption{Ablation studies on the proposed ITST. }
\label{ablation}
\end{figure}

\textbf{Constraints of Information Transport}\quad To learn information transport $\mathbf{T}$ from both translation and latency, we fuse $\mathbf{T}$ with the cross-attention for translation and introduce latency cost to constrain it. We conduct ablation studies on these two constraints, as shown in Figure \ref{ablation_a}. When removing the latency constraints, some transported weights tend to locate on the last source token (i.e., $\left<\textrm{EOS}\right>$), almost degenerating into full-sentence MT. When not fused with cross-attention, $\mathbf{T}$ cannot learn the correct information transport but simply assigns weights to the diagonal, as the latency cost around the diagonal is 0. Overall, two proposed constraints effectively help ITST learn information transport $\mathbf{T}$ from both translation and latency aspects.

\textbf{Effect of Latency Cost}\quad In Figure \ref{ablation_b}, we compare the latency cost matrix with different $\xi$. ITST is not sensitive to the setting of $\xi$ and has stable performance. More specifically, $\xi\!=\!1$ performs best since relaxing the cost of the transported weights around the diagonal allows some local reordering of information transport and meanwhile regularises the information transport. More analyses of the latency cost refer to Appendix \ref{app:latency cost}.

\textbf{Normalization of Information Transport}\quad We set the total information received by each target token to be 1 ($\sum_{j=1}^{J}T_{ij}=1$), and convert the normalization to the regular term $\mathcal{L}_{norm}$ in Eq.(\ref{eq11}) during training. To verify the normalization degree of the information transport during testing, we draw the distribution of $\sum_{j=1}^{J}T_{ij}$ in Figure \ref{ablation_c} via the boxplot. Under all latency, the information transport matrix achieves good normalization degree, where most of $\sum_{j=1}^{J}T_{ij}$ are within 1$\pm $0.05.

\subsection{Improvement on Full-sentence MT}
\label{sec:full}

\begin{table}[]
\centering
\begin{tabular}{lcc}\hlinew{1.2pt}
       \textbf{Model}            & \textbf{BLEU}  & $\Delta$ \\\hline
Transformer (Offline)        & 31.60 &       \\\hline
\textbf{Transformer + IT}     & \textbf{32.21} & \textbf{+0.61}  \\
$\;\;\;$- w/o $\mathcal{L}_{latency}$      & 31.81 & +0.21  \\
$\;\;\;$- w/o $\mathcal{L}_{norm}$      & 31.70 & +0.10  \\
$\;\;\;$- w/o $\mathcal{L}_{latency}$, $\mathcal{L}_{norm}$ & 31.62 & +0.02 \\\hlinew{1.2pt}
\end{tabular}
\caption{Full-sentence MT results on WMT15 De$\rightarrow$En (Base). `Transformer+IT': apply the proposed information transport in full-sentence MT. `w/o $\mathcal{L}_{latency}$': remove latency cost in Eq.(\ref{eq7}). `w/o $\mathcal{L}_{norm}$': remove regular term for normalization in Eq.(\ref{eq11}). }
\label{full}
\end{table}

Modeling information transport (IT) not only guides ST to decide whether to start translating, but can also directly improve translation quality since we fuse the information transport with the attention mechanism. We report the improvement that modeling information transport (IT) brings to full-sentence MT in Table \ref{full}. IT brings an improvement of 0.61 BLEU on full-sentence MT. Specifically, the latency cost matrix encourages the information transport to be near the diagonal and thereby enhances the attention around the diagonal, which is helpful for translation \cite{dyer-etal-2013-simple}. Normalization of information transport $\mathcal{L}_{norm}$ is more important since it ensures that information transport is in a legal form. When removing both latency cost and normalization, information transport is almost out of constraints, so there is little improvement.

\begin{figure}[t]
\centering
\includegraphics[width=2.3in]{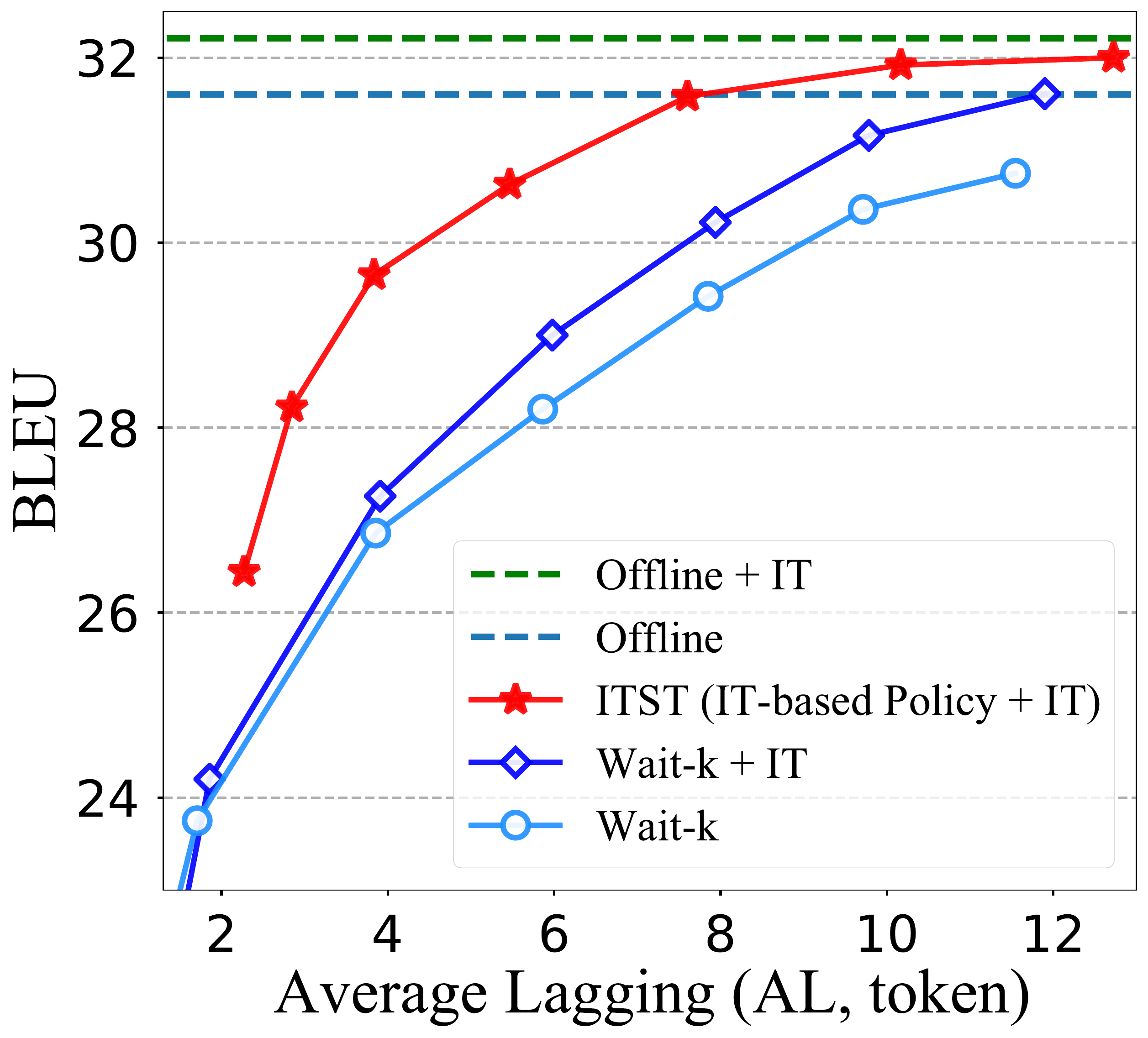}
\caption{Comparison of `Wait-k+IT' (introduce IT in `Wait-k' policy) and ITST (IT-based policy+IT), showing the improvement brought by the proposed policy.}
\label{it}
\end{figure}

Furthermore, to verify that the improvement of ITST on ST is not only due to the improvement brought by modeling IT in translation, but also due to the superiority of the proposed policy, we split ITST into modeling `IT' in translation and `IT-based policy' and compare the improvements brought by these two parts. To this end, we combine `IT' with the previous `Wait-k' to show the specific improvements brought by IT-based policy. As shown in Figure \ref{it}, when both applying IT, ITST still outperforms `Wait-k+IT', showing that more improvements of ITST are brought by the IT-based policy. More specifically, modeling IT improves the translation quality of `Wait-k' under high latency, but only has little improvement at low latency, indicating that the improvements of ITST at low latency are mainly because IT-based policy provides a more reasonable read/write policy for ST. We will in-depth evaluate the quality of read/write policy in ITST in Sec.\ref{sec:Quality of Read/Write Policy in ITST}.

\subsection{Improvement on Non-streaming Speech Translation}

\begin{table}[t]
\centering
\begin{tabular}{lc} \hlinew{1.2pt}
\textbf{Model} & \textbf{BLEU}  \\\hline
Fairseq ST \cite{wang-etal-2020-fairseq}     & 22.7           \\
ESPnet ST \cite{inaguma-etal-2020-espnet}     & 22.9           \\
AFS \cite{zhang-etal-2020-adaptive}            & 22.4           \\
DDT \cite{le-etal-2020-dual}           & 23.6           \\
RealTranS \cite{zeng-etal-2021-realtrans}     & 23.0           \\
\textbf{ITST}  & \textbf{24.4} \\\hlinew{1.2pt}
\end{tabular}
\caption{Non-streaming speech translation results on MuST-C En$\rightarrow$De.}
\label{non-streaming}
\end{table}

ITST can also be applied to non-streaming speech translation (a.k.a., offline speech translation) by removing the read/write policy. We report the performance of ITST on non-streaming speech translation in Table \ref{non-streaming}. Compare with the previous works, ITST achieved an improvement of about 1 BLEU.

\subsection{Quality of Read/Write Policy in ITST}
\label{sec:Quality of Read/Write Policy in ITST}

\begin{figure}[t]
\centering
\includegraphics[width=2.4in]{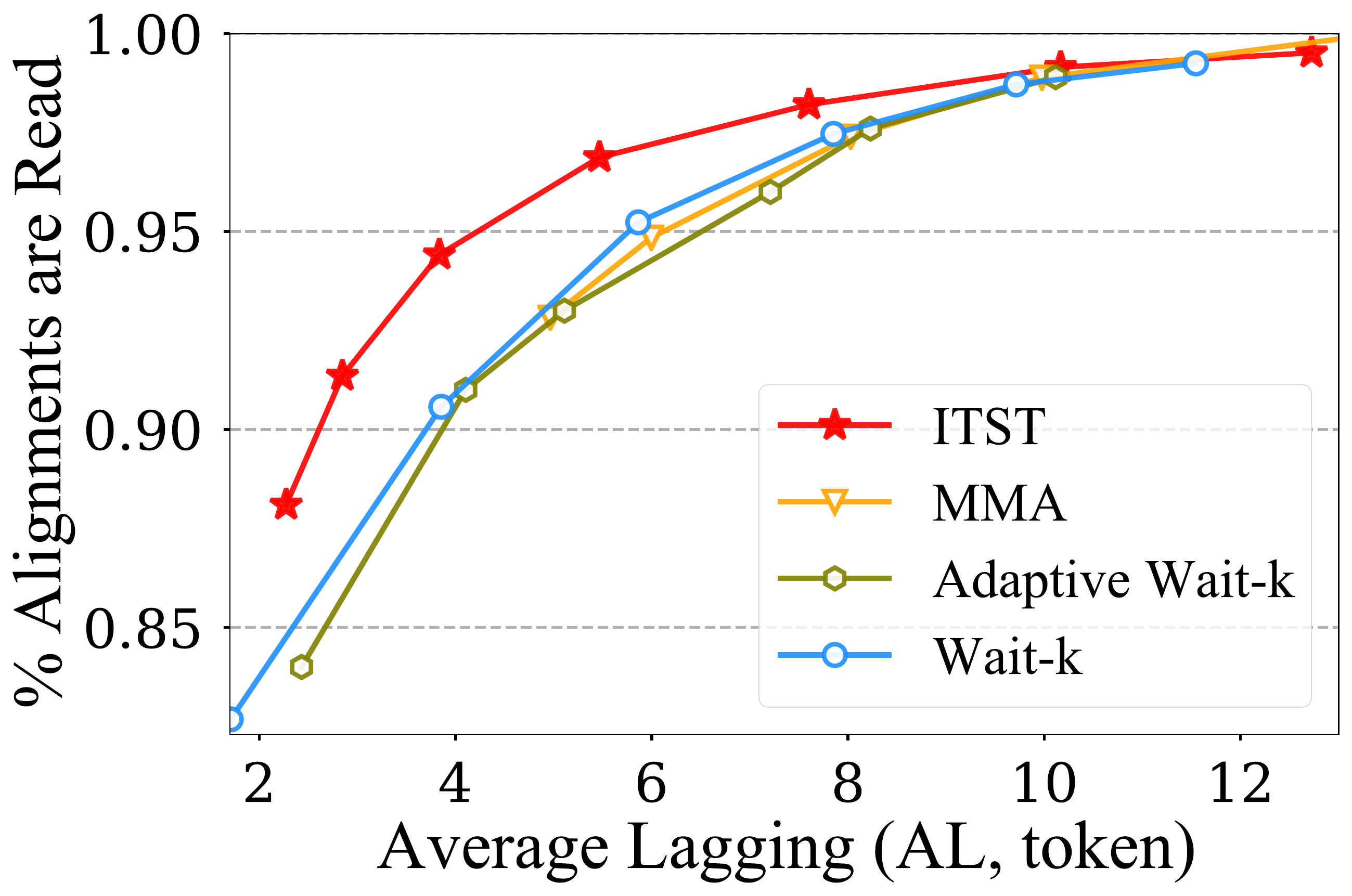}
\caption{Quality of read/write policy. We calculate the proportion of aligned source tokens received before translating (the higher ratio is better).}
\label{suff}
\end{figure}

A good read/write policy should ensure that the model translates each target token after receiving its aligned source token for translation faithfulness. To evaluate the quality of read/write policy, we calculate the proportion of the ground-truth aligned source tokens received before translating \cite{dualpath,post-eval} on RWTH\footnote{\url{https://www-i6.informatik.rwth-aachen.de/goldAlignment/}. For many-to-one alignment, we choose the last source position in the alignment.} De$\rightarrow$En alignment dataset. We denote the ground-truth aligned source position of $y_{i}$ as $a_{i}$, and denote the number of source tokens received when the read/write policy decides to translate $y_{i}$ as $g_{i}$. Then, the proportion of aligned source tokens received before translating is calculated as $\frac{1}{I}\sum_{i=1}^{I}\mathbbm{1}_{a_{i}\leq g_{i}}$, where $\mathbbm{1}_{a_{i}\leq g_{i}}$ counts the number that $a_{i}\leq g_{i}$, i.e., the number of aligned source tokens received before the read/write policy decides to translate. 

The results are shown in Figure \ref{suff}. Compared with `Waitk', `Adaptive Wait-k' and `MMA', ITST can receive more aligned source tokens before translating under the same latency. Especially under low latency, ITST can receive about 5\% more aligned tokens before translating than previous policies. Overall, the results show that ITST develops a more reasonable read/write policy that reads more aligned tokens for high translation quality and avoids unnecessary waits to keep low latency.

\subsection{Superiority of Curriculum-based Training}
\label{sec:cur}

\begin{figure}[t]
\centering
\includegraphics[width=2.4in]{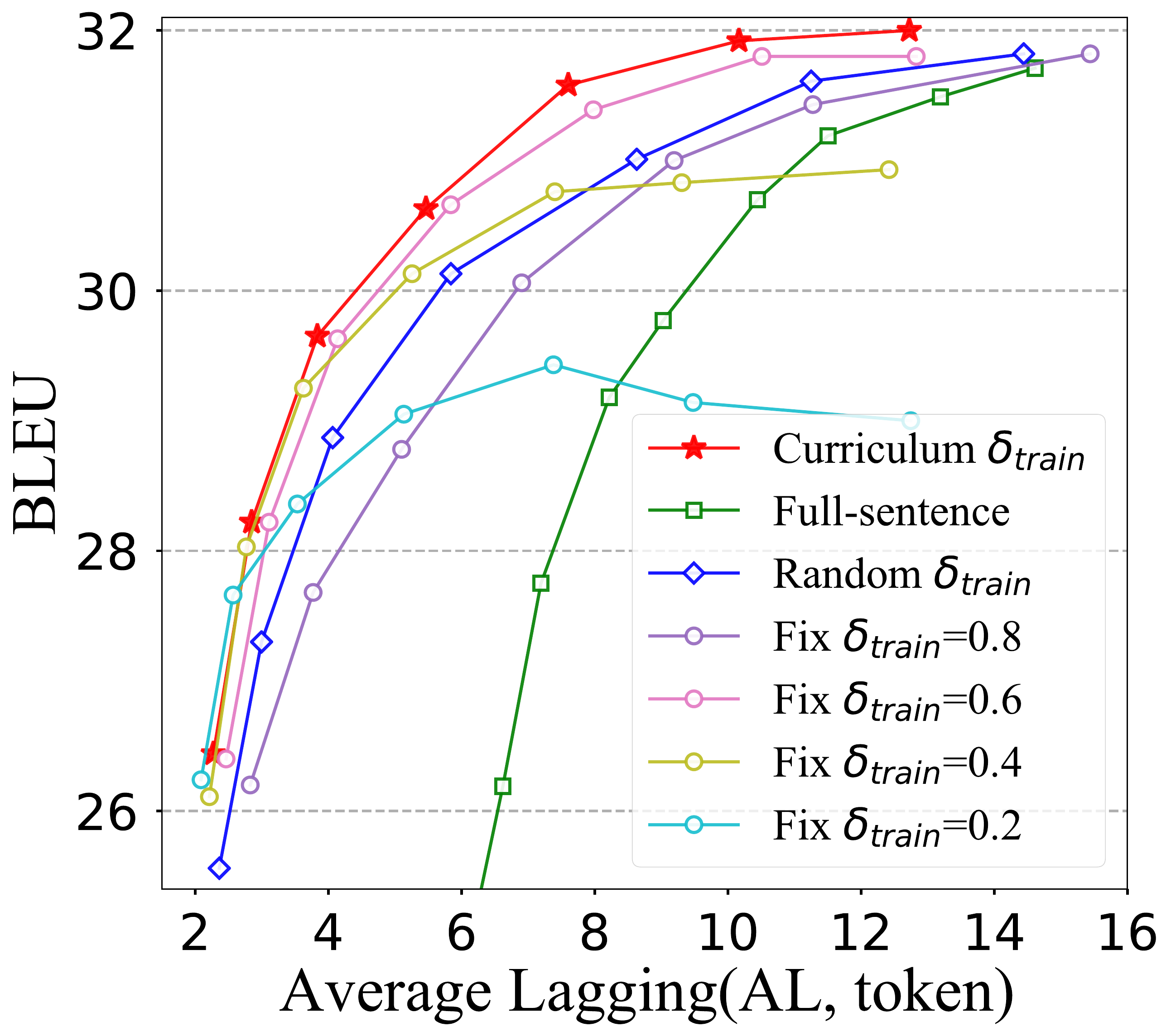}
\caption{Comparison of different training methods.}
\label{train}
\end{figure}

In Figure \ref{train}, we compare different training methods, including the proposed curriculum-based training (refer to Eq.(\ref{eq17})), using a fixed $\delta_{train}$ \cite{ma-etal-2019-stacl}, randomly sampling $\delta_{train}$ \cite{multipath} or directly applying full-sentence training \cite{Cho2016,siahbani-etal-2018-simultaneous}.

When applying full-sentence training, the obvious train-test mismatch results in poor ST performance. For fixing $\delta_{train}$, ST model can only perform well under partial latency in testing, e.g., `Fix $\delta_{train}\!=\!0.2$' performs well at low latency, but the translation quality is degraded under high latency, which is consistent with the previous conclusions \cite{ma-etal-2019-stacl,zheng-etal-2020-simultaneous}. Randomly sampling $\delta_{train}$ improves generalization under different latency, but fails to achieve the best translation quality under all latency \cite{multipath} since it ignores the correlation between different latency. In curriculum-based training, the model first learns full-sentence MT and then gradually turns to learn ST, following an easy-to-hard schedule, so it reduces the learning difficulty and thereby achieves the best translation quality under all latency.

\section{Related Work}

\textbf{Read/Write Policy}\quad Existing read/write policies fall into fixed and adaptive. For fixed policy, \citet {ma-etal-2019-stacl} proposed wait-k policy, which starts translating after receiving $k$ source tokens. \citet{multipath} enhanced wait-k policy by sampling different $k$ during training. \citet{future-guided} proposed future-guide training for wait-k policy. \citet{zhang-feng-2021-icts} proposed a char-level wait-k policy. \citet{zhang-feng-2021-universal} proposed MoE wait-k to develop a universal ST model. 

For adaptive policy, early policies used segmented translation \cite{bangalore-etal-2012-real,Cho2016,siahbani-etal-2018-simultaneous}. \citet {gu-etal-2017-learning} used reinforcement learning to train an agent. \citet{Zheng2019b} trained the policy with a rule-based READ/WRITE sequence. \citet {Zheng2019a} added a ``delay'' token to read source tokens. \citet {Arivazhagan2019} proposed MILk, predicting a Bernoulli variable to decide READ/WRITE. \citet {Ma2019a} proposed MMA to implement MILk on Transformer. \citet{miao-etal-2021-generative} proposed a generative framework to predict READ/WRITE. \citet{dualpath} proposed a dual-path method to enhance read/write policy in MMA. \citet{gma} proposed Gaussian multi-head attention to decide READ/WRITE according to alignments. ITST develops a read/write policy by modeling the translation process as information transport and taking the received information as evidence of READ/WRITE.

\textbf{Training Method of ST}\quad Early ST methods are directly trained with full-sentence MT \cite{Cho2016,siahbani-etal-2018-simultaneous}, where obvious train-test mismatching results in low translation quality. To avoid mismatching, most methods separately train multiple ST models for different latency \cite{ma-etal-2019-stacl,Ma2019a,miao-etal-2021-generative}, resulting in large computational costs. Recently, some works develop the universal ST model for different latency via randomly sampling latency in training \cite{multipath,zhang-feng-2021-universal}, but ignore the correlation between different latency. We propose a curriculum-based training for ITST to improve both efficiency and performance, which is also suitable for other read/write policies.

\section{Conclusion}
In this paper, we treat the translation as the information transport from source to target and accordingly propose information-transport-based policy (ITST). Experiments on both text-to-text and speech-to-text ST tasks show the superiority of ITST in terms of performance, training method and policy quality.

\section*{Limitations}
The room for the improvement of ITST lies in modeling information transport. Although jointly learning information transport with cross-attention is verified to be effective for ST tasks in our work, we believe that there are still some parts that can be further improved, such as using some more refined methods to analyze the contribution of source tokens to translation. However, those more refined methods also need to address the challenges such as the way of integrating into ST model, avoiding the increase in decoding time (due to the low-latency requirement of ST task), and accurately analyzing with partial source and target contents in ST. We put the exploration of such methods and challenges into our future work.

\section*{Acknowledgements}
We thank all the anonymous reviewers for their insightful and valuable comments.

\bibliography{anthology,custom}
\bibliographystyle{acl_natbib}

\clearpage
\appendix

\section{Expanded Experiments}
\label{sec:appendix}

\subsection{Why Designing Latency Cost Matrix as Diagonal Form?}
\label{app:latency cost}

\begin{figure*}[t]
\centering
\subfigure[Diagonal]{
\includegraphics[width=2in]{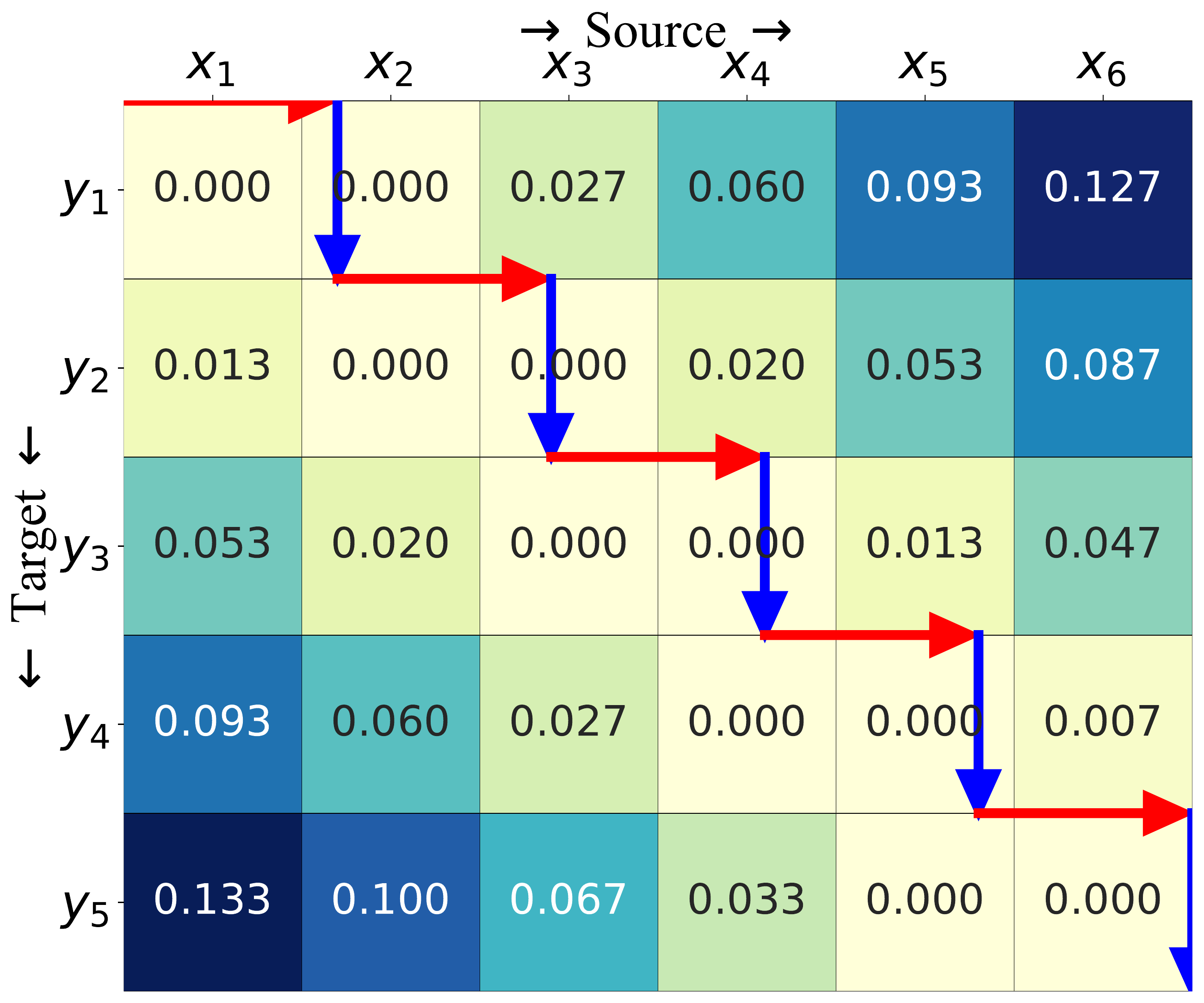} \label{cost_a}
}
\subfigure[Upper Triangular]{
\includegraphics[width=2in]{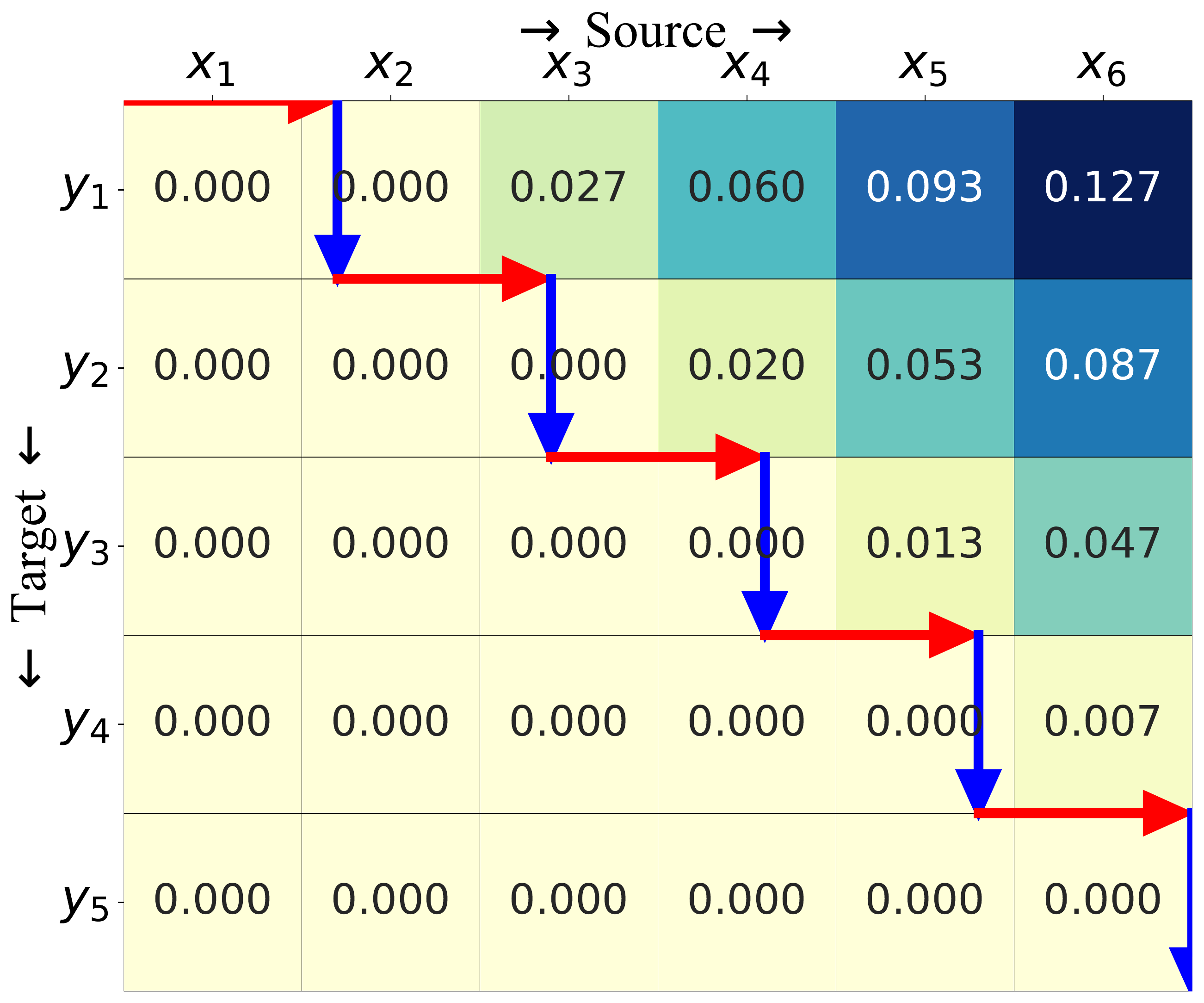} \label{cost_b}
}
\subfigure[Lower Triangular]{
\includegraphics[width=2in]{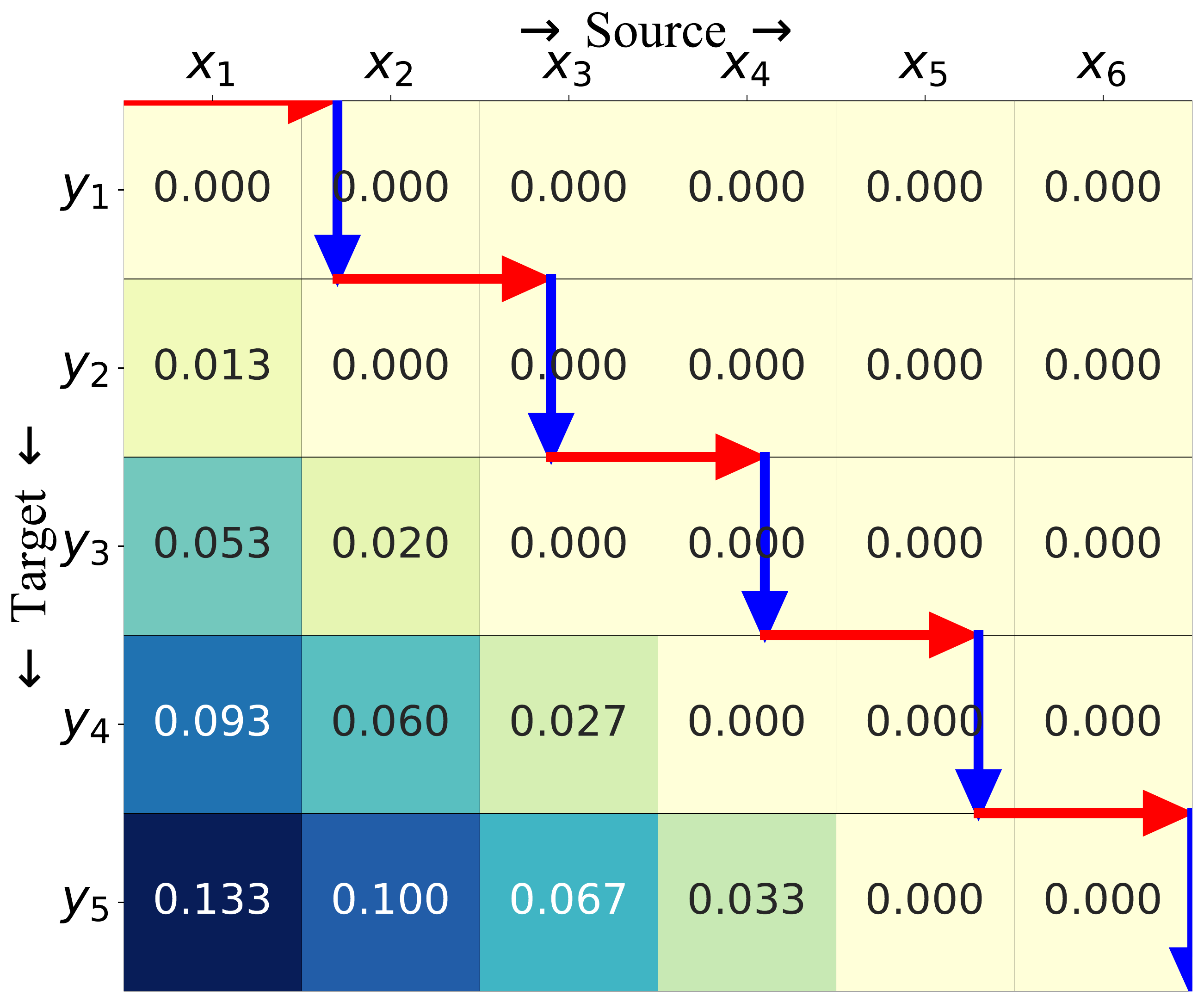} \label{cost_c}
}
\caption{Three forms of latency cost matrix. }
\label{cost}
\end{figure*}

To explore the impact of the latency cost matrix form on ST performance, we compared three different forms of the latency cost matrix:
\begin{itemize}
\setlength{\itemsep}{0pt}
\setlength{\parsep}{0pt}
\setlength{\parskip}{0pt}
    \item \textbf{Diagonal}: The information transport far away from the diagonal cost more, as shown in Figure \ref{cost_a}.
    \item \textbf{Upper Triangular}: Only constrain the transported weights from source tokens that lag behind. The information transport below the diagonal cost 0 (i.e., transport from $x_{j}$ to $y_{i}$ cost 0 when $j\!<\!i\!\times\! J/I$), and the rest are the same as `Diagonal', as shown in Figure \ref{cost_b}.
    \item \textbf{Lower Triangular}: Only constrain the transported weights from source tokens in the front. The information transport above the diagonal cost 0 (i.e., transport from $x_{j}$ to $y_{i}$ cost 0 when $j\!>\!i\!\times\! J/I$), and the rest are the same as `Diagonal', as shown in Figure \ref{cost_c}.
\end{itemize}

\begin{figure}[t]
\centering
\includegraphics[width=2.4in]{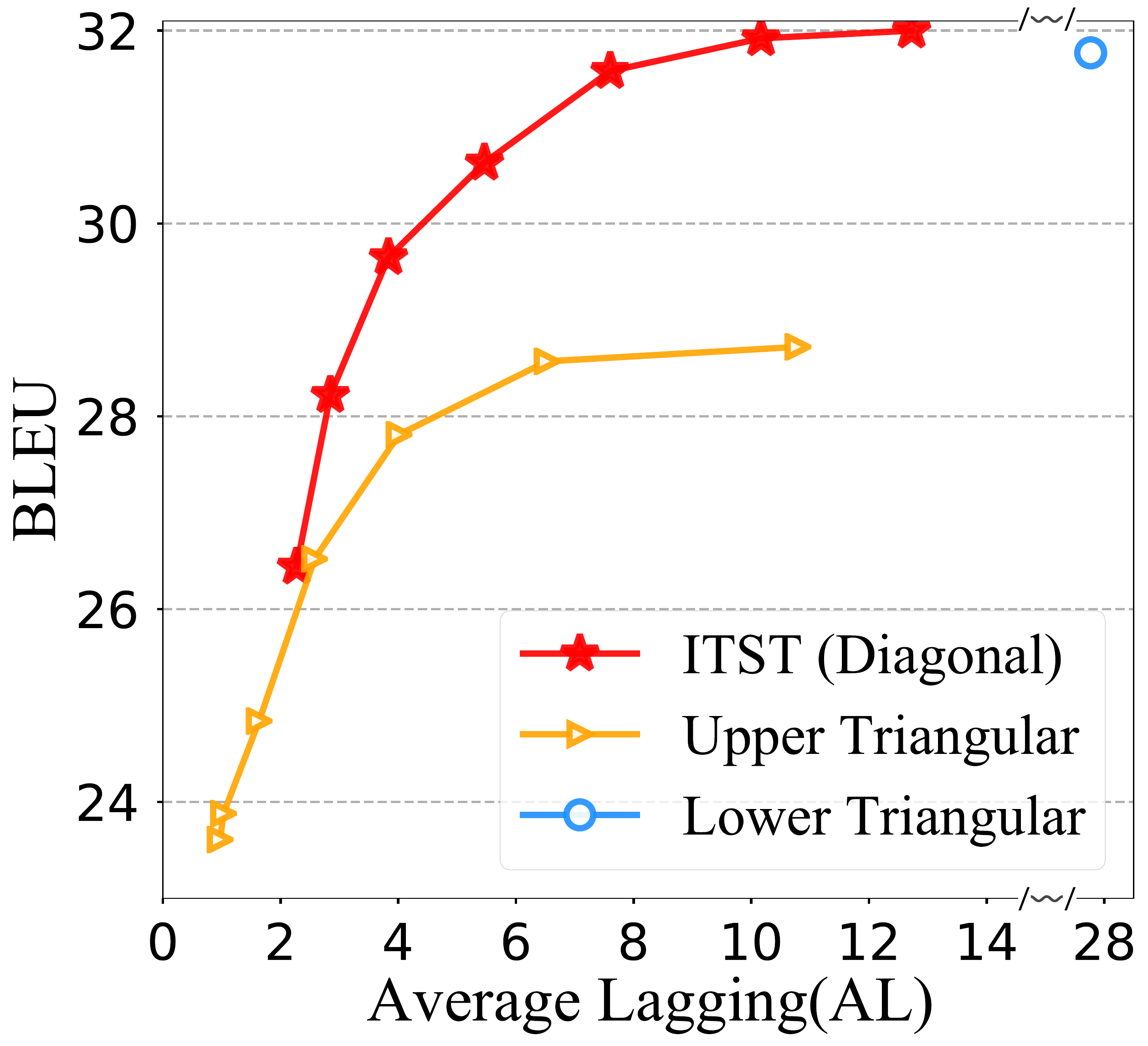}
\caption{Comparison of various forms of latency cost matrix $\mathbf{C}$.}
\label{cost_type}
\end{figure}

We show the results of these three latency cost matrices in Figure \ref{cost_type}. In `Upper Triangular', the information provided by the front source tokens is almost unconstrained, so that more transported weights will be transported from the front tokens. Accordingly, the accumulated source information will exceed the threshold much earlier, resulting in much prematurely outputting and lower translation quality. While in `Lower Triangular', due to the lack of constraints on weights transported from later source tokens, some transported weights will tend to locate on the last source token (i.e., $\left<\textrm{EOS}\right>$), resulting in higher latency.
In contrast, the proposed latency cost in `Diagonal' form avoids too much information weight being transported from source tokens that are located too early or too late compared to the position of the current target token, so the model can perform ST at an almost constant speed and thus perform better.

\subsection{Case Study}

\begin{figure*}[t]
\centering
\includegraphics[width=6.3in]{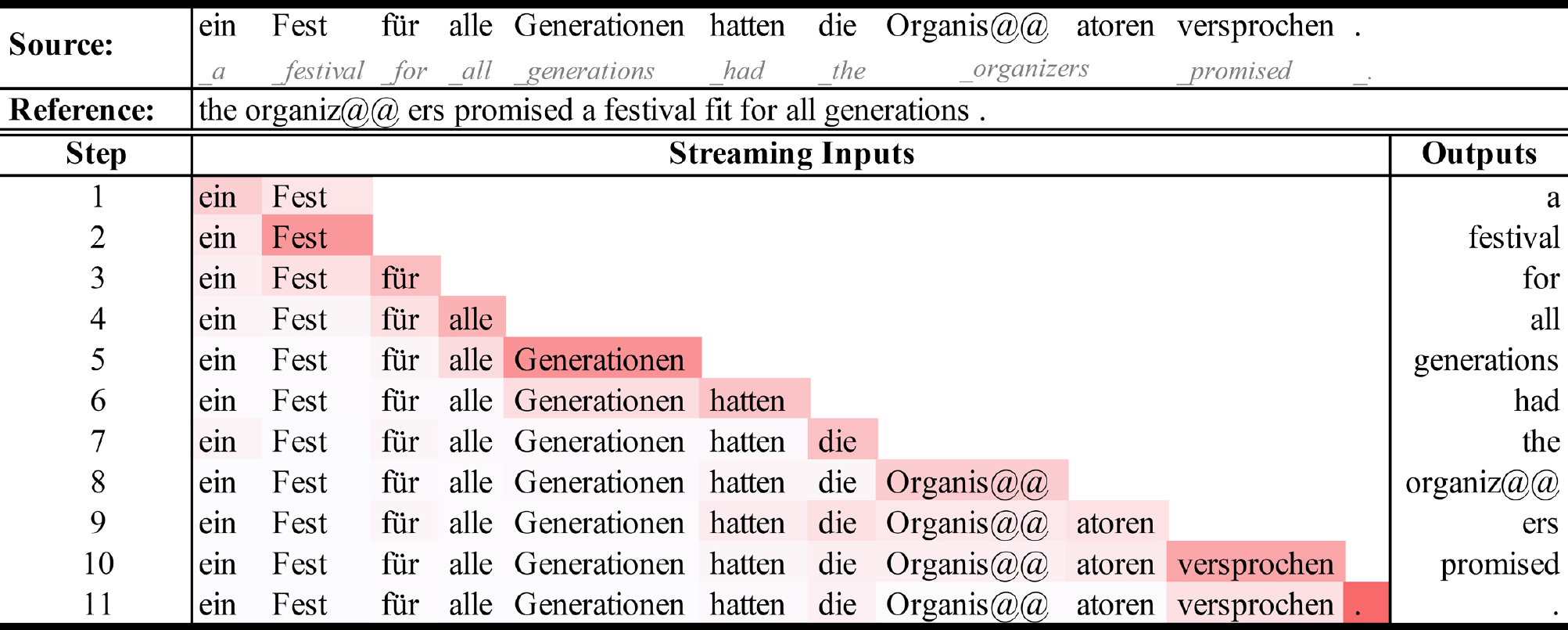}
\caption{Case No.1741 in WMT15 De$\rightarrow$En test set, evaluated with  $\delta=0.2$. The background color represents the information transport from source token to target token, where the darker color indicate larger transported weight.}
\label{case0.2}
\end{figure*}
\begin{figure*}[t]
\centering
\includegraphics[width=6.3in]{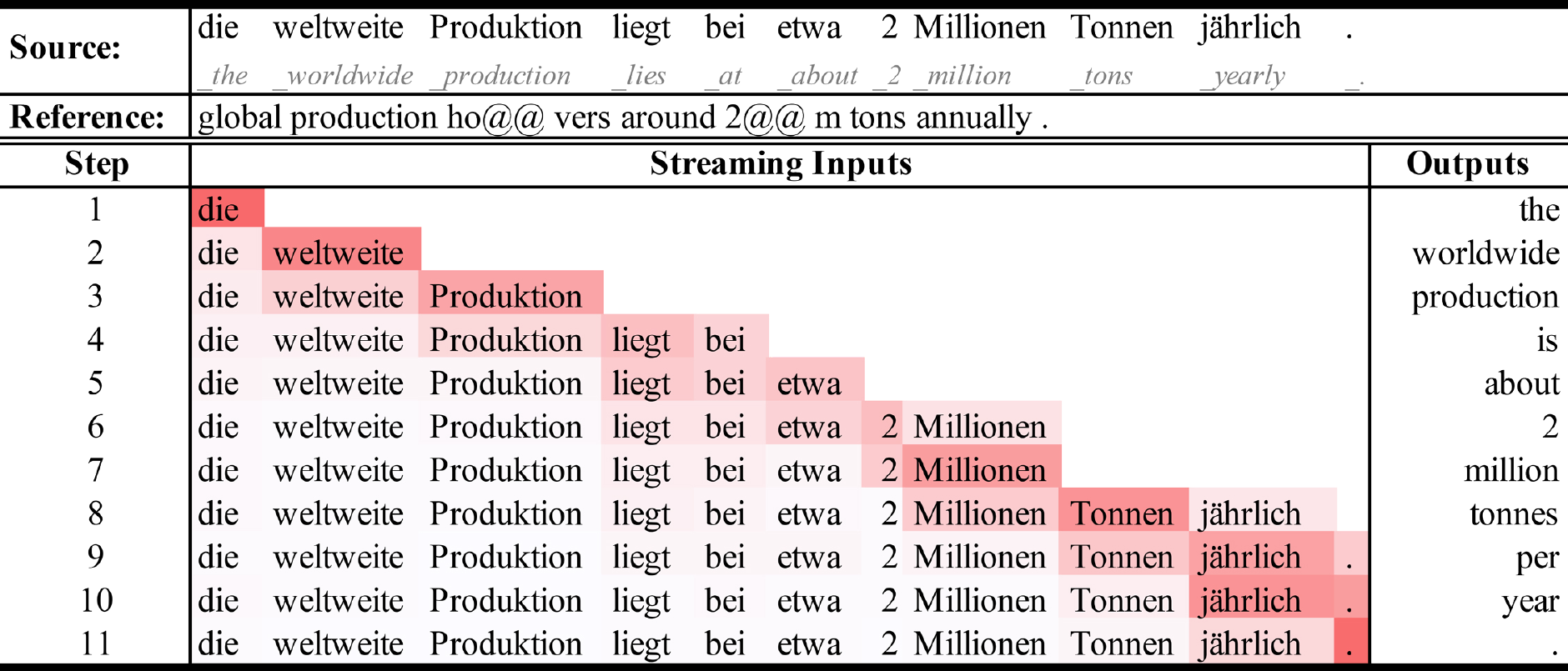}
\caption{Case No.1272 in WMT15 De$\rightarrow$En test set, evaluated with  $\delta=0.5$. The background color represents the information transport from source token to target token, where the darker color indicate larger transported weight.}
\label{case0.5}
\end{figure*}
\begin{figure*}[t]
\centering
\includegraphics[width=6.3in]{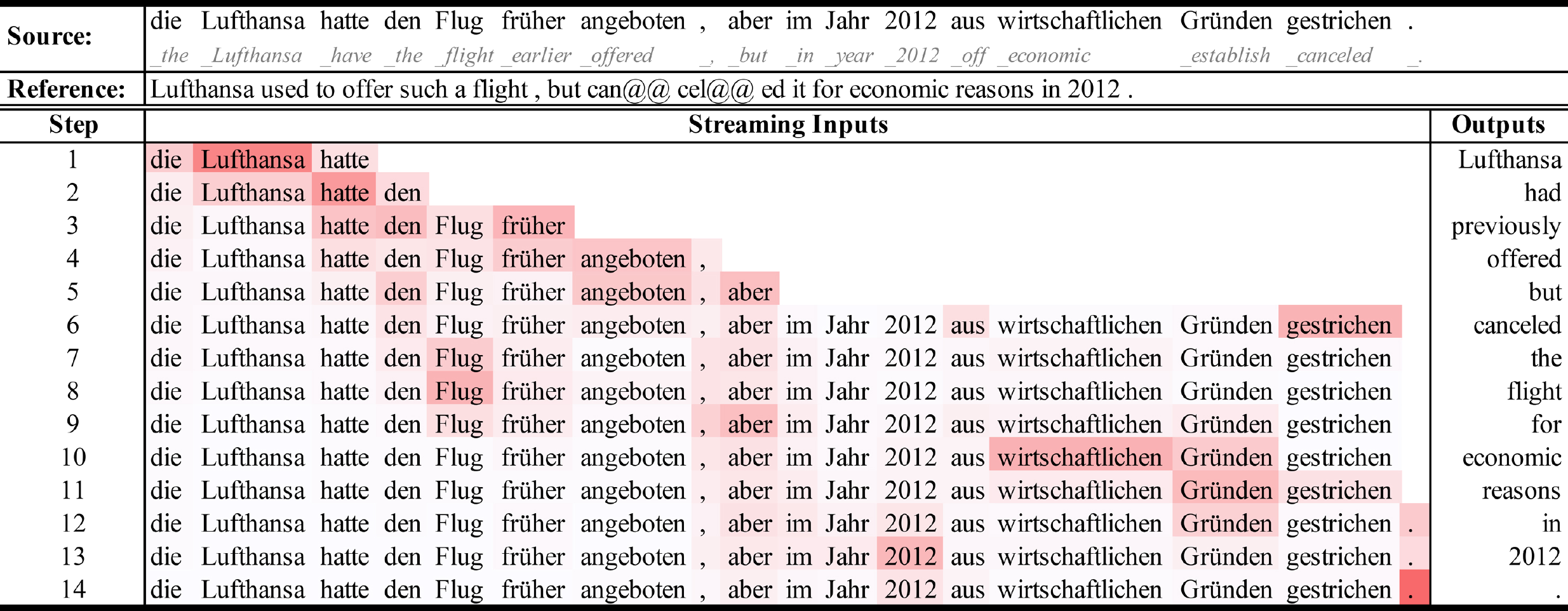}
\caption{Case No.1675 in WMT15 De$\rightarrow$En test set, evaluated with  $\delta=0.8$. The background color represents the information transport from source token to target token, where the darker color indicate larger transported weight.}
\label{case0.8}
\end{figure*}

\begin{figure*}[t]
\centering
\includegraphics[width=6.3in]{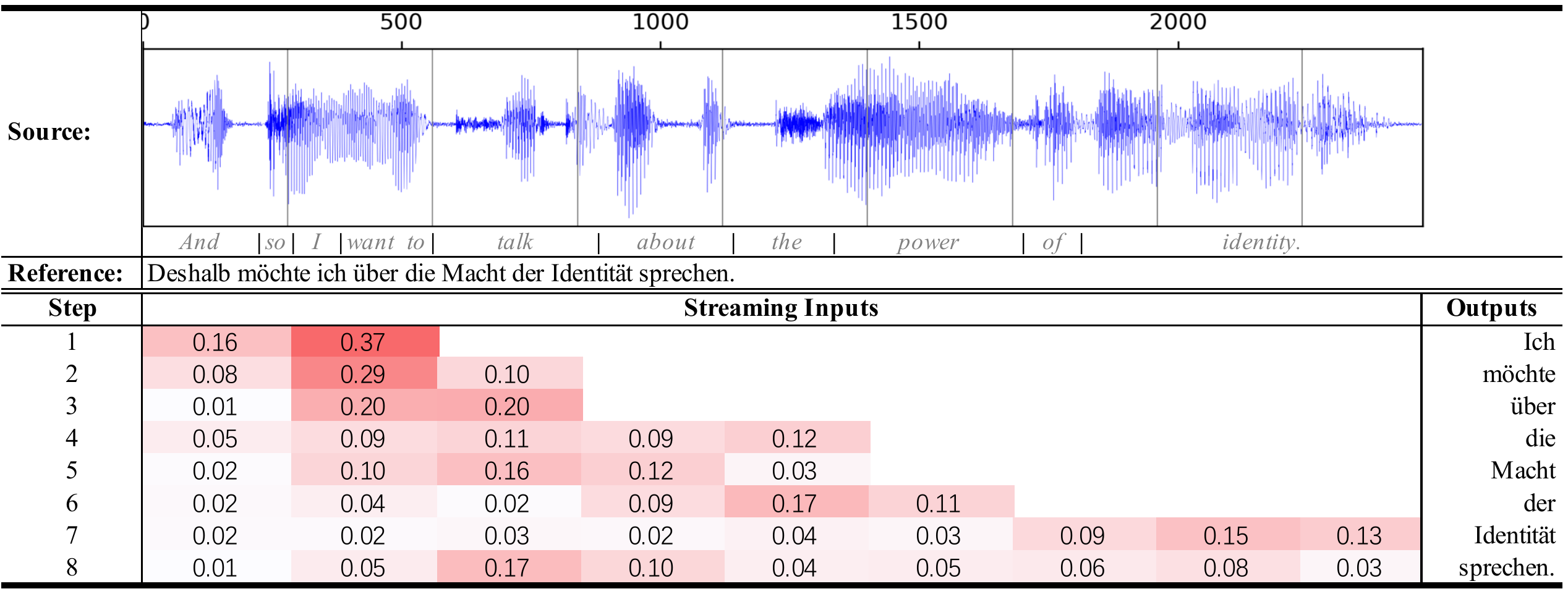}
\caption{Case ted\_1378\_201 in MuST-C En$\rightarrow$De test set, evaluated with $\delta=0.4$ and make decision of READ/WRITE every 7 source token (280$ms$).}
\label{s2t_case0.4-1378}
\end{figure*}

\begin{figure*}[t]
\centering
\includegraphics[width=6.3in]{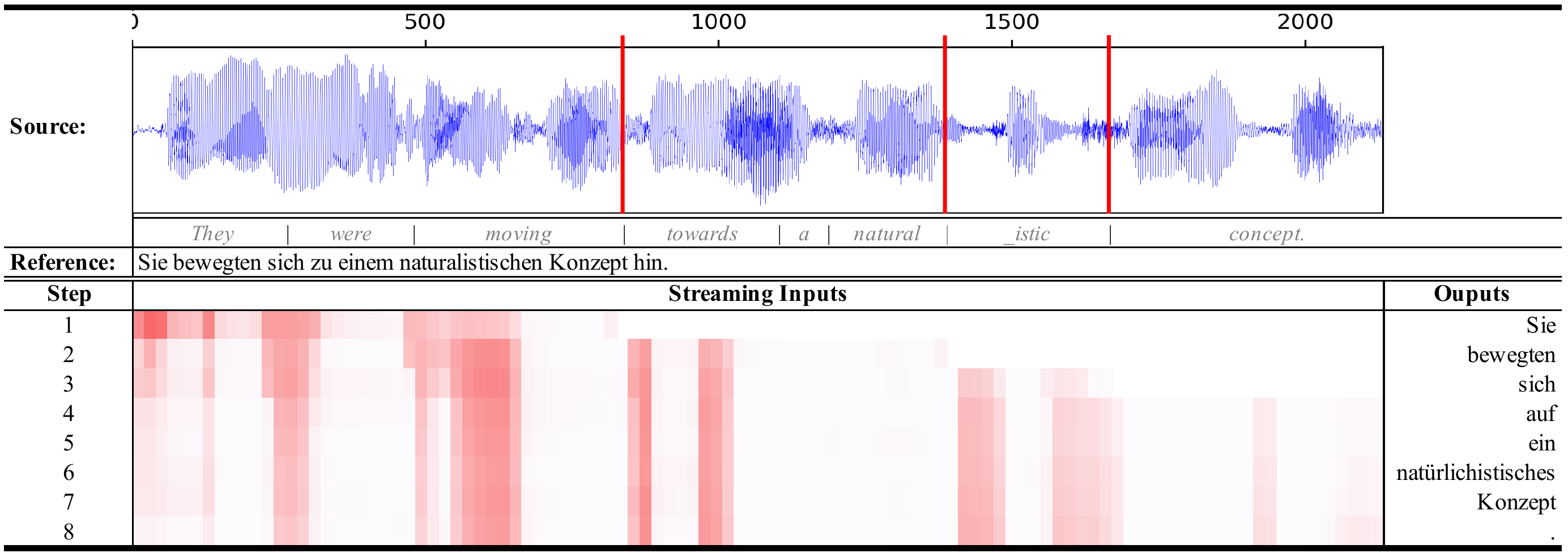}
\caption{Case ted\_1166\_102 in MuST-C En$\rightarrow$De test set, evaluated with $\delta=0.7$ and flexible pre-decision on each speech frame.}
\label{s2t_case0.7-1166}
\end{figure*}

We conduct case studies to explore the characteristics of ITST, especially the information transport. As shown in Figure \ref{case0.2}, \ref{case0.5}, \ref{case0.8}, \ref{s2t_case0.4-1378} and \ref{s2t_case0.7-1166}, we visualized the process of simultaneous translation step by step, where the background color of the source tokens represents the transported information weight to the current target token in ITST. Note that the transported information weight is not normalized, especially when the source is incomplete, which is described in Sec.\ref{sec:Learning of Information Transport}.

\textbf{Text-to-text ST (low latency)}\quad As shown in Figure \ref{case0.2}, ITST can accurately predict the transported information weight, where the corresponding source token often provides more information for the current target token. Besides, this case has a serious word order reversal between reference and source (e.g., `\textit{Organisatoren}' locates at the back of the source, but the corresponding `\textit{organizers}' is at the beginning of the reference.), which is more challenging for ST. Under a small threshold $\delta=0.2$, ITST can learn to generate a semantically-correct translation in a monotonic order, owing to the proposed latency cost matrix in a diagonal form.

\textbf{Text-to-text ST (middle latency)}\quad As shown in Figure \ref{case0.5}, as $\delta$ increases, ITST receives some nearby tokens after reading the source token with the largest transported weight. This enables ITST to obtain richer source information and meanwhile efficiently handle the many-to-one alignments (e.g., `\textit{liegt bei}' is translated to `\textit{is}' as a whole).

\textbf{Text-to-text ST (high latency)}\quad As shown in Figure \ref{case0.8}, higher thresholds allow the model to wait until the corresponding source token and then start translating, e.g., ITST waits until the corresponding `\textit{gestrichen}' before translating `\textit{canceled}'. Meanwhile, the information transport predicted by ITST exhibits a strong locally-reordering ability to satisfy more complex alignments between the target and source.

\textbf{Speech-to-text ST (fixed pre-decision)}\quad As shown in Figure \ref{s2t_case0.4-1378}, ITST also performs well on speech-to-text ST with fixed pre-decision, where the information transport matrix can effectively capture the information transport from the speech segment (280$ms$) to the target token.

\textbf{Speech-to-text ST (flexible pre-decision)}\quad As shown in Figure \ref{s2t_case0.7-1166}, ITST can effectively model the information transport from each speech frame to the target token, thereby identifying which frames are more important. Therefore, ITST can perform segmentation at the frame with less information transport, and thereby flexibly divide the source speech into multiple speech segments with complete meaning, achieving a more reasonable read/write policy.

\section{Numerical Results}
\label{sec:Numerical Results}
\textbf{More Latency Metrics}\quad Besides Average Lagging (AL) \cite{ma-etal-2019-stacl}, we also use Consecutive Wait (CW) \cite{gu-etal-2017-learning}, Average Proportion (AP) \cite{Cho2016} and Differentiable Average Lagging (DAL) \cite{Arivazhagan2019} to evaluate the latency of the ST model, all of which are calculated based on $g_{i}$. For text-to-text ST, $g_{i}$ records the number of source tokens received when translating $y_{i}$. For speech-to-text ST, $g_{i}$ records the speech duration ($ms$) read in when translating $y_{i}$. Besides, we also use computation aware latency (denotes as CW-CA, AP-CA, AL-CA and DAL-CA) \cite{ma-etal-2020-simulmt} for speech-to-text ST to consider the computational time of the model, where $g_{i}$ records the absolute moment when translating $y_{i}$. All computation aware latency are evaluated on 1 NVIDIA 3090 GPU with $\text{batch-size}=1$. 
The calculations of latency metrics are as follows.

\textbf{Consecutive Wait (CW)} \cite{gu-etal-2017-learning} evaluates the average number of source tokens waited between two target tokens. Given $g_{i}$, CW is calculated as:
\begin{gather}
    \mathrm{CW}=\frac{\sum_{i=1}^{\left | \mathbf{y} \right |} (g_{i}-g_{i-1})}{\sum_{i=1}^{\left | \mathbf{y} \right |}\mathbbm{1}_{g_{i}-g_{i-1}>0}},
\end{gather}
where $\mathbbm{1}_{g_{i}-g_{i-1}>0}=1$ counts the number of $g_{i}-g_{i-1}>0$.

\textbf{Average Proportion (AP)} \cite{Cho2016} measures the proportion of the received source tokens before translating. Given $g_{i}$, AP is calculated as:
\begin{gather}
    \mathrm{AP}=\frac{1}{\left | \mathbf{x} \right | \left | \mathbf{y} \right |}\sum_{i=1}^{\left | \mathbf{y} \right |} g_{i}.
\end{gather}

\textbf{Average Lagging (AL)} \cite{ma-etal-2019-stacl} evaluates the number of tokens that the outputs lags behind the inputs. Given $g_{i}$, AL is calculated as:
\begin{align}
    \mathrm{AL}=&\;\frac{1}{\tau }\sum_{i=1}^{\tau}g_{i}-\frac{i-1}{\left | \mathbf{y} \right |/\left | \mathbf{x} \right |},\\
\mathrm{where}\;\; \tau =&\;\underset{i}{\mathrm{argmax}}\left ( g_{i}= \left | \mathbf{x} \right |\right ).
\end{align}
$\left | \mathbf{x} \right |$ and $\left | \mathbf{y} \right |$ are the length of the source sequence and target sequence respectively.

\textbf{Differentiable Average Lagging (DAL)} \cite{Arivazhagan2019} is a differentiable version of average lagging. Given $g_{i}$, DAL is calculated as:
\begin{align}
g^{'}_{i}=&\;\left\{\begin{matrix}
g_{i} & i=1\\ 
 \mathrm{max}\left (g_{i},g^{'}_{i-1}+ \frac{\left | \mathbf{x} \right |}{\left | \mathbf{y} \right |} \right )& i>1
\end{matrix}\right., \\
    \mathrm{DAL}=&\;\frac{1}{\left | \mathbf{y} \right | }\sum\limits_{i=1}^{\left | \mathbf{y} \right |}g^{'}_{i}-\frac{i-1}{\left | \mathbf{x} \right |/\left | \mathbf{y} \right |}.
\end{align}

\textbf{Numerical Results}\quad Table \ref{envi_t2t}, \ref{deenbase_t2t}, \ref{deenbig_t2t}, \ref{ende_s2t}, \ref{enes_s2t} and \ref{ende_s2t_flexible} report the numerical results of all systems in our experiments, evaluated with BLEU and SacreBLEU for translation quality and CW, AP, AL and DAL for latency.

\begin{table*}[t]
\centering
\small
\begin{tabular}{cccccc} \hlinew{1.2pt}
\multicolumn{6}{c}{\textbf{IWSLT15 English$\rightarrow$Vietnamese (Small)}}                           \\ \hline
\multicolumn{6}{c}{\textit{\textbf{Offline}}}                       \\
           & CW    & AP   & AL    & DAL   & BLEU  \\
           & 22.08 & 1.00 & 22.08 & 22.08 & 28.91 \\ \hline
\multicolumn{6}{c}{\textit{\textbf{Wait-k}}}                        \\
$k$          & CW    & AP   & AL    & DAL   & BLEU  \\
1          & 1.00  & 0.63 & 3.03  & 3.54  & 25.21 \\
3          & 1.17  & 0.71 & 4.80  & 5.42  & 27.65 \\
5          & 1.46  & 0.78 & 6.46  & 7.06  & 28.34 \\
7          & 1.96  & 0.83 & 8.21  & 8.79  & 28.60 \\
9          & 2.73  & 0.88 & 9.92  & 10.51 & 28.69 \\ \hline
\multicolumn{6}{c}{\textit{\textbf{Multipath Wait-k}}}              \\
$k$          & CW    & AP   & AL    & DAL   & BLEU  \\
1          & 1.01  & 0.63 & 3.06  & 3.61  & 26.23 \\
3          & 1.17  & 0.71 & 4.66  & 5.20  & 28.21 \\
5          & 1.46  & 0.78 & 6.38  & 6.94  & 28.56 \\
7          & 1.96  & 1.96 & 8.13  & 8.69  & 28.62 \\
9          & 2.73  & 0.87 & 9.80  & 10.34 & 28.52 \\ \hline
\multicolumn{6}{c}{\textit{\textbf{Adaptive Wait-k}}}               \\
( $\rho_{1}$, $\rho_{9}$ )   & CW    & AP   & AL    & DAL   & BLEU  \\
(0.02, 0.00)   & 1.05  & 0.63 & 2.98  & 3.64  & 25.69 \\
(0.04, 0.00)   & 1.19  & 0.63 & 3.07  & 4.06  & 26.05 \\
(0.05, 0.00)   & 1.27  & 1.27 & 3.14  & 4.30  & 26.33 \\
(0.10, 0.00)    & 1.97  & 0.68 & 4.08  & 6.05  & 27.80 \\
(0.10, 0.05) & 2.36  & 0.71 & 4.77  & 7.11  & 28.46 \\
(0.20, 0.00)    & 2.73  & 0.78 & 6.56  & 8.34  & 28.73 \\
(0.30, 0.20)  & 3.39  & 0.86 & 9.42  & 10.42 & 28.80 \\ \hline
\multicolumn{6}{c}{\textit{\textbf{MoE Wait-k}}}                    \\
$k$          & CW    & AP   & AL    & DAL   & BLEU  \\
1          & 1.00  & 0.63 & 3.19  & 3.76  & 26.56 \\
3          & 1.17  & 0.71 & 4.70  & 5.42  & 28.43 \\
5          & 1.46  & 0.78 & 6.43  & 7.14  & 28.73 \\
7          & 1.97  & 0.83 & 8.19  & 8.88  & 28.81 \\
9          & 2.73  & 0.87 & 9.86  & 10.39 & 28.88 \\ \hline
\multicolumn{6}{c}{\textit{\textbf{MMA}}}                           \\
$\lambda$     & CW    & AP   & AL    & DAL   & BLEU  \\
0.4        & 1.03  & 0.58 & 2.68  & 3.46  & 27.73 \\
 0.3        & 1.09  & 0.59 & 2.98  & 3.81  & 27.90 \\
0.2        & 1.15  & 0.63 & 3.57  & 4.44  & 28.47 \\
0.1        & 1.31  & 0.67 & 4.63  & 5.65  & 28.42 \\
0.04       & 1.64  & 0.70 & 5.44  & 6.57  & 28.33 \\
0.02       & 2.01  & 0.76 & 7.09  & 8.29  & 28.28 \\ \hline
\multicolumn{6}{c}{\textit{\textbf{ITST}}}                          \\
$\delta$      & CW    & AP   & AL    & DAL   & BLEU  \\
0.1        & 1.18  & 0.68 & 3.95  & 5.04  & 28.56 \\
0.2        & 2.08  & 0.72 & 4.55  & 8.59  & 28.68 \\
0.3        & 4.24  & 0.80 & 6.10  & 13.26 & 28.81 \\
0.4        & 6.61  & 0.88 & 8.31  & 16.61 & 28.82 \\
0.5        & 9.01  & 0.92 & 10.75 & 18.73 & 28.89 \\ \hlinew{1.2pt}
\end{tabular}
\caption{Numerical results of text-to-text ST on IWSLT15 En$\rightarrow$Vi with Transformer-small. As the raw data from \url{nlp.stanford.edu/projects/nmt/} is tokenized, we only report BLEU for IWSLT15 En$\rightarrow$Vi.}
\label{envi_t2t}
\end{table*}

\begin{table*}[t]
\centering
\small
\begin{tabular}{ccccccc}
\hlinew{1.2pt}
\multicolumn{7}{c}{\textbf{WMT15 German$\rightarrow$English (Base)}}            \\ \hline
\multicolumn{7}{c}{\textit{\textbf{Offline}}}                                   \\
           & CW    & AP   & AL    & DAL   & BLEU  & SacreBLEU \\
           & 27.77 & 1.00 & 27.77 & 27.77 & 31.60 & 30.21     \\ \hline
\multicolumn{7}{c}{\textit{\textbf{Wait-k}}}                                    \\
$k$          & CW    & AP   & AL    & DAL   & BLEU  & SacreBLEU \\
1          & 1.17  & 0.52 & 0.02  & 1.84  & 17.61 & 16.75     \\
3          & 1.23  & 0.59 & 1.71  & 3.33  & 23.75 & 22.80     \\
5          & 1.37  & 0.66 & 3.85  & 5.20  & 26.86 & 25.83     \\
7          & 1.70  & 0.73 & 5.86  & 7.12  & 28.20 & 27.15     \\
9          & 2.17  & 0.78 & 7.85  & 9.01  & 29.42 & 27.99     \\
11         & 2.78  & 0.82 & 9.71  & 10.79 & 30.36 & 29.27     \\
13         & 3.56  & 0.86 & 11.55 & 12.49 & 30.75 & 29.65     \\ \hline
\multicolumn{7}{c}{\textit{\textbf{Multipath Wait-k}}}                          \\
$k$          & CW    & AP   & AL    & DAL   & BLEU  & SacreBLEU \\
1          & 1.27  & 0.50 & -0.49 & 1.60  & 19.51 & 18.62     \\
3          & 1.27  & 0.58 & 1.56  & 3.29  & 24.11 & 23.12     \\
5          & 1.39  & 0.66 & 3.71  & 5.18  & 26.85 & 25.82     \\
7          & 1.71  & 0.73 & 5.78  & 7.12  & 28.34 & 27.31     \\
9          & 2.17  & 0.78 & 7.84  & 8.98  & 29.39 & 28.34     \\
11         & 2.78  & 0.82 & 9.73  & 10.79 & 30.02 & 28.98     \\
13         & 3.56  & 0.86 & 11.50 & 12.49 & 30.25 & 29.20     \\ \hline
\multicolumn{7}{c}{\textit{\textbf{Adaptive Wait-k}}}                           \\
( $\rho_{1}$, $\rho_{13}$ )   & CW    & AP   & AL    & DAL   & BLEU  & SacreBLEU \\
(0.02, 0.00)   & 1.54  & 0.54 & 0.83  & 3.27  & 20.29 & 19.31     \\
(0.04, 0.00)   & 2.07  & 0.56 & 1.40  & 4.59  & 22.34 & 21.43     \\
(0.05, 0.00)   & 2.28  & 0.58 & 1.90  & 5.25  & 23.56 & 22.64     \\
(0.06, 0.00)   & 2.58  & 0.60 & 2.43  & 5.99  & 24.59 & 23.65     \\
(0.07, 0.00)   & 2.79  & 0.62 & 2.94  & 6.57  & 25.96 & 24.99     \\
(0.09, 0.00)   & 3.25  & 0.66 & 4.10  & 7.78  & 27.44 & 26.42     \\
(0.10, 0.00)    & 3.45  & 0.68 & 4.66  & 8.31  & 27.88 & 26.83     \\
(0.10, 0.01) & 3.68  & 0.70 & 5.11  & 8.84  & 28.29 & 27.24     \\
(0.10, 0.03) & 4.13  & 0.72 & 6.09  & 9.87  & 28.91 & 27.87     \\
(0.10, 0.05)  & 4.48  & 0.75 & 7.21  & 10.72 & 29.73 & 28.63     \\
(0.20, 0.00)    & 4.02  & 0.78 & 8.23  & 10.92 & 30.10 & 28.96     \\
(0.02, 0.05) & 4.75  & 0.82 & 10.12 & 12.35 & 30.76 & 29.66     \\
(0.20, 0.10)  & 4.68  & 0.85 & 11.55 & 12.98 & 30.78 & 29.68     \\
(0.30, 0.20)  & 4.16  & 0.86 & 12.18 & 13.09 & 30.74 & 29.65     \\ \hline
\multicolumn{7}{c}{\textit{\textbf{MoE Wait-k}}}                                \\
$k$          & CW    & AP   & AL    & DAL   & BLEU  & SacreBLEU \\
1          & 1.49  & 0.49 & -0.32 & 1.69  & 21.43 & 20.31     \\
3          & 1.26  & 0.59 & 1.79  & 3.30  & 25.81 & 24.70     \\
5          & 1.37  & 0.66 & 3.88  & 5.18  & 28.34 & 27.31     \\
7          & 1.69  & 0.73 & 5.94  & 7.12  & 29.71 & 28.65     \\
9          & 2.17  & 0.78 & 7.86  & 8.99  & 30.61 & 29.50     \\
11         & 2.78  & 0.82 & 9.73  & 10.78 & 30.89 & 29.76     \\
13         & 3.56  & 0.86 & 11.53 & 12.48 & 31.08 & 29.98     \\ \hline
\multicolumn{7}{c}{\textit{\textbf{MMA}}}                                       \\
$\lambda$     & CW    & AP   & AL    & DAL   & BLEU  & SacreBLEU \\
0.4        & 2.35  & 0.68 & 4.97  & 7.51  & 28.66 & 27.69     \\
0.3        & 2.64  & 0.72 & 6.00  & 9.30  & 29.11 & 28.19     \\
0.25       & 3.35  & 0.78 & 8.03  & 12.28 & 28.92 & 28.02     \\
0.2        & 4.03  & 0.83 & 9.98  & 14.86 & 28.18 & 27.29     \\
0.1        & 14.88 & 0.97 & 13.25 & 19.48 & 27.47 & 26.60     \\ \hline
\multicolumn{7}{c}{\textit{\textbf{GSiMT}}}                                     \\
$\zeta$         & CW    & AP   & AL    & DAL   & BLEU  & SacreBLEU \\
4          & -     & -    & 3.64  & -     & 28.82 & -         \\
5          & -     & -    & 4.45  & -     & 29.50 & -         \\
6          & -     & -    & 5.13  & -     & 29.78 & -         \\
7          & -     & -    & 6.24  & -     & 29.63 & -         \\ \hline
\multicolumn{7}{c}{\textit{\textbf{ITST}}}                                      \\
$\delta$      & CW    & AP   & AL    & DAL   & BLEU  & SacreBLEU \\
0.2        & 1.43  & 0.59 & 2.27  & 3.87  & 26.44 & 25.17     \\
0.3        & 1.70  & 0.61 & 2.85  & 4.86  & 28.22 & 26.94     \\
0.4        & 2.16  & 0.65 & 3.83  & 6.61  & 29.65 & 28.58     \\
0.5        & 3.18  & 0.71 & 5.47  & 10.16 & 30.63 & 29.51     \\
0.6        & 4.63  & 0.78 & 7.60  & 14.24 & 31.58 & 30.46     \\
0.7        & 7.04  & 0.86 & 10.17 & 19.17 & 31.92 & 30.74     \\
0.8        & 9.78  & 0.91 & 12.72 & 22.52 & 32.00 & 30.84  \\ \hlinew{1.2pt}
\end{tabular}
\caption{Numerical results of text-to-text ST on WMT15 De$\rightarrow$En with Transformer-Base.}
\label{deenbase_t2t}
\end{table*}

\begin{table*}[t]
\small
\centering
\begin{tabular}{ccccccc}\hlinew{1.2pt}
\multicolumn{7}{c}{\textbf{WMT15 German$\rightarrow$English (Big)}}             \\\hline
\multicolumn{7}{c}{\textit{\textbf{Offline}}}                                   \\
           & CW    & AP   & AL    & DAL   & BLEU  & SacreBLEU \\
           & 27.77 & 1.00 & 27.77 & 27.77 & 32.94 & 31.35     \\ \hline
\multicolumn{7}{c}{\textit{\textbf{Wait-k}}}                                    \\
$k$          & CW    & AP   & AL    & DAL   & BLEU  & SacreBLEU \\
1          & 1.16  & 0.52 & 0.25  & 1.82  & 19.13 & 18.13     \\
3          & 1.20  & 0.60 & 2.23  & 3.41  & 25.45 & 24.30     \\
5          & 1.36  & 0.67 & 4.00  & 5.23  & 28.67 & 27.52     \\
7          & 1.70  & 0.73 & 5.97  & 7.17  & 30.12 & 28.97     \\
9          & 2.17  & 0.78 & 7.95  & 9.03  & 31.46 & 30.27     \\
11         & 2.79  & 0.82 & 9.75  & 10.82 & 31.83 & 30.63     \\
13         & 3.56  & 0.86 & 11.59 & 12.51 & 32.08 & 30.95     \\ \hline
\multicolumn{7}{c}{\textit{\textbf{Multipath Wait-k}}}                          \\
$k$          & CW    & AP   & AL    & DAL   & BLEU  & SacreBLEU \\
1          & 1.23  & 0.51 & -0.19 & 1.79  & 20.56 & 19.45     \\
3          & 1.26  & 0.59 & 1.73  & 3.36  & 25.45 & 24.43     \\
5          & 1.39  & 0.66 & 3.82  & 5.24  & 28.58 & 27.55     \\
7          & 1.71  & 0.73 & 5.89  & 7.16  & 30.13 & 29.04     \\
9          & 2.17  & 0.78 & 7.88  & 9.02  & 31.23 & 30.14     \\
11         & 2.78  & 0.82 & 9.77  & 10.81 & 31.52 & 30.37     \\
13         & 3.56  & 0.86 & 11.58 & 12.51 & 32.02 & 30.83     \\ \hline
\multicolumn{7}{c}{\textit{\textbf{Adaptive Wait-k}}}                           \\
( $\rho_{1}$, $\rho_{13}$ )   & CW    & AP   & AL    & DAL   & BLEU  & SacreBLEU \\
(0.02, 0.00)   & 1.42  & 0.54 & 0.99  & 3.00  & 20.50 & 19.50     \\
(0.04, 0.00)   & 1.86  & 0.56 & 1.37  & 4.22  & 22.62 & 21.55     \\
(0.05, 0.00)   & 2.10  & 0.57 & 1.69  & 4.81  & 23.77 & 22.71     \\
(0.06, 0.00)   & 2.36  & 0.59 & 2.23  & 5.54  & 25.43 & 24.38     \\
(0.07, 0.00)   & 2.58  & 0.61 & 2.70  & 6.14  & 27.06 & 26.01     \\
(0.08, 0.00)   & 2.84  & 0.63 & 3.17  & 6.75  & 27.96 & 26.94     \\
(0.09, 0.00)   & 3.08  & 0.65 & 3.72  & 7.33  & 28.92 & 27.80     \\
(0.10, 0.00)    & 3.28  & 0.67 & 4.28  & 7.88  & 29.90 & 28.82     \\
(0.10, 0.03) & 3.95  & 0.71 & 5.59  & 9.43  & 30.97 & 29.83     \\
(0.10, 0.05) & 4.36  & 0.74 & 6.70  & 10.41 & 31.30 & 30.11     \\
(0.20, 0.00)    & 3.90  & 0.78 & 8.09  & 10.80 & 32.38 & 31.18     \\
(0.20, 0.05) & 4.78  & 0.82 & 10.00 & 12.35 & 32.46 & 31.27     \\
(0.30, 0.20)  & 4.16  & 0.86 & 12.19 & 13.11 & 32.24 & 31.06     \\ \hline
\multicolumn{7}{c}{\textit{\textbf{MoE Wait-k}}}                                \\
$k$          & CW    & AP   & AL    & DAL   & BLEU  & SacreBLEU \\
1          & 1.41  & 0.51 & 0.16  & 1.79  & 21.76 & 20.52     \\
3          & 1.28  & 0.59 & 2.03  & 3.37  & 26.51 & 25.30     \\
5          & 1.37  & 0.67 & 4.03  & 5.22  & 29.33 & 28.11     \\
7          & 1.70  & 0.73 & 5.95  & 7.14  & 30.66 & 29.45     \\
9          & 2.17  & 0.78 & 7.86  & 8.99  & 30.61 & 29.50     \\
11         & 2.78  & 0.82 & 9.73  & 10.78 & 30.89 & 29.76     \\
13         & 3.56  & 0.86 & 11.53 & 12.48 & 31.08 & 29.98     \\ \hline
\multicolumn{7}{c}{\textit{\textbf{MMA}}}                                       \\
$\lambda$     & CW    & AP   & AL    & DAL   & BLEU  & SacreBLEU \\
1          & 1,69  & 0.56 & 3.00  & 4.03  & 26.10 & 25.10     \\
0.75       & 1.66  & 0.58 & 3.40  & 4.46  & 26.50 & 25.50     \\
0.5        & 1.69  & 0.59 & 3.69  & 4.83  & 27.70 & 26.70     \\
0.4        & 1.70  & 0.59 & 3.75  & 4.90  & 29.20 & 28.24     \\
0.3        & 1.82  & 0.60 & 4.18  & 5.35  & 30.30 & 29.26     \\
0.27       & 2.37  & 0.71 & 5.91  & 8.27  & 30.88 & 29.88     \\
0.25       & 2.62  & 0.75 & 7.02  & 9.88  & 31.04 & 30.00     \\
0.2        & 3.21  & 0.79 & 8.75  & 12.60 & 31.08 & 30.04     \\ \hline
\multicolumn{7}{c}{\textit{\textbf{ITST}}}                                      \\
$\delta$      & CW    & AP   & AL    & DAL   & BLEU  & SacreBLEU \\
0.2        & 1.33  & 0.58 & 1.89  & 3.62  & 25.90 & 24.73     \\
0.3        & 1.48  & 0.60 & 2.44  & 4.21  & 27.51 & 26.75     \\
0.4        & 1.70  & 0.62 & 2.99  & 4.91  & 29.35 & 28.52     \\
0.5        & 2.04  & 0.66 & 4.09  & 6.42  & 30.83 & 29.99     \\
0.6        & 2.98  & 0.72 & 6.07  & 9.95  & 31.90 & 31.05     \\
0.7        & 4.59  & 0.81 & 8.60  & 15.03 & 32.85 & 32.02     \\
0.8        & 7.23  & 0.89 & 11.37 & 20.05 & 32.90 & 32.09 \\\hlinew{1.2pt}
\end{tabular}
\caption{Numerical results of text-to-text ST on WMT15 De$\rightarrow$En, with Transformer-Big.}
\label{deenbig_t2t}
\end{table*}

\begin{table*}[]
\centering
\small
\begin{tabular}{cccccccccc}\hlinew{1.2pt}
\multicolumn{10}{c}{\textbf{MuST-C English$\rightarrow$German}}                                                \\\hline
\multicolumn{10}{c}{\textit{\textbf{Offline}}}                                            \\
       & CW      & CW-CA   & AP   & AP-CA & AL      & AL-CA   & DAL     & DAL-CA  & SacreBLEU  \\
       & 5654.72 & 6793.53 & 1.00 & 1.26  & 5654.72 & 6437.17 & 5654.72 & 6439.38 & 16.24 \\\hline
\multicolumn{10}{c}{\textit{\textbf{Wait-k}}}                                             \\
$k$       & CW      & CW-CA   & AP   & AP-CA & AL      & AL-CA   & DAL     & DAL-CA  & SacreBLEU  \\
1      & 539.80  & 656.95  & 0.49 & 0.66  & 798.33  & 1091.60 & 898.82  & 1167.53 & 4.84  \\
3      & 510.89  & 625.65  & 0.67 & 0.88  & 1265.59 & 1651.38 & 1419.25 & 1827.51 & 9.98  \\
5      & 624.23  & 794.96  & 0.77 & 1.18  & 1716.74 & 2206.40 & 1914.74 & 2466.57 & 12.36 \\
7      & 798.04  & 986.47  & 0.83 & 1.13  & 2154.41 & 2661.35 & 2377.99 & 2952.11 & 14.16 \\
9      & 1019.29 & 1256.31 & 0.87 & 1.17  & 2548.04 & 3095.09 & 2764.27 & 3379.53 & 14.76 \\
11     & 1248.96 & 1533.52 & 0.90 & 1.20  & 2889.36 & 3465.67 & 3102.10 & 3742.53 & 15.19 \\
13     & 1490.41 & 1830.52 & 0.92 & 1.21  & 3198.89 & 3817.19 & 3398.02 & 4070.75 & 15.58 \\\hline
\multicolumn{10}{c}{\textit{\textbf{Multipath Wait-k}}}                                   \\
$k$       & CW      & CW-CA   & AP   & AP-CA & AL      & AL-CA   & DAL     & DAL-CA  & SacreBLEU  \\
1      & 383.28  & 485.09  & 0.63 & 0.93  & 778.01  & 1265.81 & 1078.94 & 1683.84 & 11.66 \\
3      & 472.26  & 601.06  & 0.73 & 1.05  & 1270.75 & 1828.58 & 1556.94 & 2265.29 & 13.69 \\
5      & 606.17  & 769.77  & 0.80 & 1.12  & 1718.82 & 2327.06 & 1989.23 & 2736.57 & 14.63 \\
7      & 795.19  & 989.01  & 0.85 & 1.19  & 2138.85 & 2809.33 & 2408.08 & 3217.05 & 15.32 \\
9      & 1011.12 & 1275.68 & 0.88 & 1.23  & 2530.30 & 3255.44 & 2785.97 & 3648.38 & 15.55 \\
11     & 1245.04 & 1563.39 & 0.91 & 1.27  & 2878.54 & 3642.61 & 3117.31 & 4002.51 & 15.63 \\
13     & 1492.89 & 1852.43 & 0.93 & 1.31  & 3195.34 & 4024.85 & 3418.42 & 4372.89 & 15.78 \\\hline
\multicolumn{10}{c}{\textit{\textbf{MoE Wait-k}}}                                         \\
$k$       & CW      & CW-CA   & AP   & AP-CA & AL      & AL-CA   & DAL     & DAL-CA  & SacreBLEU  \\
1      & 390.56  & 490.69  & 0.63 & 0.90  & 798.55  & 1294.49 & 1081.45 & 1711.97 & 11.82 \\
3      & 478.09  & 615.60  & 0.73 & 1.00  & 1276.90 & 1855.19 & 1541.88 & 2271.15 & 14.02 \\
5      & 611.17  & 771.08  & 0.80 & 1.08  & 1738.15 & 2373.90 & 2001.28 & 2796.13 & 15.23 \\
7      & 799.59  & 1010.62 & 0.85 & 1.14  & 2157.08 & 2839.02 & 2415.62 & 3253.87 & 15.83 \\
9      & 1015.63 & 1278.19 & 0.88 & 1.17  & 2542.59 & 3255.72 & 2781.72 & 3631.79 & 16.05 \\
11     & 1248.64 & 1550.77 & 0.91 & 1.22  & 2892.69 & 3661.99 & 3111.84 & 4007.84 & 16.16 \\
13     & 1502.20 & 1879.50 & 0.92 & 1.24  & 3206.46 & 4008.34 & 3415.37 & 4339.51 & 16.19 \\\hline
\multicolumn{10}{c}{\textit{\textbf{MMA}}}                                                \\
$\lambda$ & CW      & CW-CA   & AP   & AP-CA & AL      & AL-CA   & DAL     & DAL-CA  & SacreBLEU  \\
0.1    & 689.14  & 861.53  & 0.55 & 0.76  & 994.26  & 1424.52 & 1172.73 & 1496.73 & 3.90  \\
0.08   & 799.81  & 1016.64 & 0.73 & 0.98  & 1530.85 & 2100.06 & 1841.43 & 2475.70 & 13.06 \\
0.04   & 1518.12 & 1927.83 & 0.85 & 1.24  & 2313.61 & 3010.78 & 2903.04 & 3607.73 & 15.24 \\
0.03   & 1953.59 & 2475.41 & 0.89 & 1.27  & 2732.72 & 3482.06 & 3347.95 & 4048.94 & 15.47 \\
0.02   & 2342.68 & 2959.23 & 0.91 & 1.29  & 3072.43 & 3845.77 & 3705.91 & 4411.64 & 15.40 \\
0.01   & 3752.20 & 4751.53 & 0.97 & 1.29  & 4225.67 & 5087.84 & 4698.99 & 5453.24 & 16.07 \\\hline
\multicolumn{10}{c}{\textit{\textbf{ITST}}}                                               \\
$\delta$  & CW      & CW-CA   & AP   & AP-CA & AL      & AL-CA   & DAL     & DAL-CA  & SacreBLEU  \\
0.2    & 452.35  & 561.12  & 0.71 & 0.91  & 1083.33 & 1551.50 & 1476.65 & 2082.17 & 14.40 \\
0.3    & 464.82  & 571.65  & 0.74 & 0.93  & 1207.42 & 1633.16 & 1593.02 & 2150.07 & 14.81 \\
0.4    & 500.59  & 605.87  & 0.77 & 0.96  & 1386.12 & 1831.00 & 1761.58 & 2337.25 & 15.15 \\
0.5    & 795.99  & 663.46  & 0.80 & 1.01  & 1595.69 & 2093.80 & 1964.17 & 2607.17 & 15.41 \\
0.6    & 658.69  & 795.99  & 0.83 & 1.04  & 1911.04 & 2410.43 & 2265.77 & 2898.05 & 15.68 \\
0.7    & 929.34  & 1118.60 & 0.88 & 1.09  & 2430.46 & 2954.02 & 2827.25 & 3500.93 & 16.12 \\
0.75   & 1216.87 & 1467.56 & 0.91 & 1.13  & 2797.97 & 3340.56 & 3323.25 & 4051.59 & 16.17 \\
0.8    & 1644.32 & 2036.56 & 0.94 & 1.17  & 3277.64 & 3839.61 & 3999.19 & 4793.56 & 16.10 \\
0.85   & 2394.93 & 2878.30 & 0.96 & 1.19  & 3877.90 & 4445.11 & 4636.44 & 5410.85 & 16.08 \\
0.9    & 3338.74 & 3995.25 & 0.98 & 1.20  & 4494.97 & 5109.77 & 5121.08 & 5889.54 & 16.17 \\\hlinew{1.2pt}
\end{tabular}
\caption{Numerical results of speech-to-text ST on MuST-C En$\rightarrow$De with fixed pre-decision of 280$ms$.}
\label{ende_s2t}
\end{table*}

\begin{table*}[]
\small
\centering
\begin{tabular}{cccccccccc}\hlinew{1.2pt}
\multicolumn{10}{c}{\textbf{MuST-C English$\rightarrow$Spanish}}                                                     \\\hline
\multicolumn{10}{c}{\textit{\textbf{Offline}}}                                                 \\
       & CW      & CW-CA   & AP      & AP-CA   & AL      & AL-CA   & DAL     & DAL-CA  & SacreBLEU  \\
       & 5998.00 & 7215.87 & 5998.00 & 6855.94 & 1.00    & 1.25    & 5998.00 & 6857.45 & 21.47 \\\hline
\multicolumn{10}{c}{\textit{\textbf{Wait-k}}}                                                  \\
$k$      & CW      & CW-CA   & AP      & AP-CA   & AL      & AL-CA   & DAL     & DAL-CA  & SacreBLEU  \\
1      & 574.62  & 701.31  & 0.47    & 0.68    & 774.45  & 1091.67 & 933.21  & 1220.33 & 6.08  \\
3      & 509.26  & 625.77  & 0.65    & 0.89    & 1163.85 & 1585.48 & 1401.56 & 1813.85 & 13.04 \\
5      & 612.59  & 747.98  & 0.75    & 0.97    & 1612.71 & 2092.52 & 1891.23 & 2375.40 & 16.48 \\
7      & 768.38  & 955.68  & 0.81    & 1.16    & 2034.91 & 2584.13 & 2342.82 & 2918.98 & 18.38 \\
9      & 967.99  & 1201.17 & 0.86    & 1.20    & 2431.37 & 3051.55 & 2760.91 & 3407.34 & 19.59 \\
11     & 1188.29 & 1465.64 & 0.89    & 1.22    & 2800.15 & 3449.49 & 3122.15 & 3794.29 & 19.95 \\
13     & 1435.10 & 1772.10 & 0.91    & 1.28    & 3135.88 & 3822.66 & 3445.35 & 4148.07 & 20.11 \\\hline
\multicolumn{10}{c}{\textit{\textbf{Multipath Wait-k}}}                                        \\
$k$       & CW      & CW-CA   & AP      & AP-CA   & AL      & AL-CA   & DAL     & DAL-CA  & SacreBLEU  \\
1      & 364.78  & 483.22  & 0.62    & 1.03    & 605.28  & 1100.93 & 1015.05 & 1588.68 & 13.87 \\
3      & 447.59  & 586.81  & 0.71    & 1.12    & 1108.27 & 1678.73 & 1494.53 & 2148.42 & 16.85 \\
5      & 572.25  & 734.35  & 0.78    & 1.19    & 1555.46 & 2147.86 & 1944.58 & 2590.77 & 17.78 \\
7      & 752.35  & 960.32  & 0.83    & 1.24    & 1998.25 & 2648.71 & 2368.27 & 3058.17 & 18.37 \\
9      & 963.75  & 1207.48 & 0.87    & 1.26    & 2406.62 & 3061.86 & 2766.38 & 3443.00 & 18.65 \\
11     & 1188.85 & 1487.81 & 0.89    & 1.30    & 2778.67 & 3478.02 & 3120.53 & 3838.07 & 18.79 \\
13     & 1434.24 & 1786.39 & 0.92    & 1.33    & 3122.23 & 3850.29 & 3438.91 & 4181.58 & 19.07 \\\hline
\multicolumn{10}{c}{\textit{\textbf{MoE Wait-k}}}                                              \\
$k$       & CW      & CW-CA   & AP      & AP-CA   & AL      & AL-CA   & DAL     & DAL-CA  & SacreBLEU  \\
1      & 390.59  & 476.90  & 0.59    & 0.76    & 674.79  & 1061.06 & 1023.30 & 1442.18 & 14.04 \\
3      & 469.55  & 586.83  & 0.70    & 0.89    & 1152.56 & 1660.50 & 1497.06 & 2053.02 & 17.65 \\
5      & 592.52  & 727.84  & 0.77    & 0.96    & 1615.52 & 2137.10 & 1959.20 & 2508.71 & 19.45 \\
7      & 768.52  & 973.23  & 0.82    & 1.06    & 2050.33 & 2731.01 & 2391.71 & 3133.42 & 20.45 \\
9      & 977.89  & 1239.99 & 0.86    & 1.11    & 2463.56 & 3191.17 & 2789.95 & 3574.23 & 21.09 \\
11     & 1202.05 & 1510.93 & 0.89    & 1.13    & 2824.93 & 3566.36 & 3134.90 & 3912.16 & 21.43 \\
13     & 1453.84 & 1799.11 & 0.91    & 1.14    & 3161.43 & 3855.42 & 3455.77 & 4162.40 & 21.66 \\\hline
\multicolumn{10}{c}{\textit{\textbf{MMA}}}                                                     \\
$\lambda$ & CW      & CW-CA   & AP      & AP-CA   & AL      & AL-CA   & DAL     & DAL-CA  & SacreBLEU  \\
0.1    & 642.49  & 796.14  & 0.60    & 0.81    & 988.50  & 1503.06 & 1272.07 & 1666.14 & 10.48 \\
0.08   & 687.79  & 859.30  & 0.74    & 1.04    & 1290.83 & 2038.35 & 1877.45 & 2823.46 & 14.98 \\
0.04   & 1077.21 & 1371.43 & 0.80    & 1.10    & 1704.66 & 2498.00 & 2387.05 & 3249.45 & 19.20 \\
0.03   & 1231.47 & 1585.13 & 0.82    & 1.09    & 1878.53 & 2663.61 & 2597.54 & 3432.40 & 19.33 \\
0.02   & 1393.97 & 1798.61 & 0.84    & 1.14    & 2104.14 & 2944.19 & 2844.50 & 3744.12 & 19.48 \\
0.01   & 1837.53 & 2317.88 & 0.88    & 1.18    & 2646.39 & 3526.19 & 3399.28 & 4305.26 & 19.77 \\
0.008  & 2101.61 & 2640.52 & 0.89    & 1.17    & 2811.79 & 3678.48 & 3614.39 & 4435.34 & 19.82 \\
0.002  & 3082.29 & 3876.83 & 0.96    & 1.25    & 4038.70 & 4966.84 & 4638.21 & 5550.06 & 20.54 \\\hline
\multicolumn{10}{c}{\textit{\textbf{ITST}}}                                                    \\
$\delta$  & CW      & CW-CA   & AP      & AP-CA   & AL      & AL-CA   & DAL     & DAL-CA  & SacreBLEU  \\
0.2    & 455.84  & 555.33  & 0.69    & 0.87    & 960.49  & 1413.42 & 1452.41 & 2006.21 & 17.77 \\
0.3    & 476.70  & 577.35  & 0.74    & 0.91    & 1152.53 & 1611.10 & 1653.25 & 2198.59 & 18.38 \\
0.4    & 510.98  & 635.66  & 0.77    & 0.97    & 1351.47 & 1907.32 & 1843.40 & 2510.41 & 18.71 \\
0.5    & 585.07  & 735.21  & 0.81    & 1.03    & 1620.54 & 2227.51 & 2112.38 & 2826.43 & 19.11 \\
0.6    & 708.80  & 883.25  & 0.84    & 1.06    & 1964.43 & 2594.40 & 2431.00 & 3151.37 & 19.77 \\
0.7    & 889.71  & 1111.49 & 0.88    & 1.10    & 2380.75 & 3057.35 & 2824.10 & 3589.36 & 20.13 \\
0.75   & 1020.53 & 1273.01 & 0.89    & 1.12    & 2642.81 & 3332.45 & 3073.15 & 3860.18 & 20.46 \\
0.8    & 1227.30 & 1521.41 & 0.91    & 1.14    & 2979.87 & 3665.20 & 3453.46 & 4250.35 & 20.75 \\
0.85   & 1583.26 & 1972.53 & 0.94    & 1.18    & 3433.96 & 4134.71 & 4002.54 & 4887.81 & 20.48 \\
0.9    & 2124.43 & 2645.45 & 0.96    & 1.21    & 3982.66 & 4701.74 & 4662.57 & 5610.53 & 20.64 \\\hlinew{1.2pt}
\end{tabular}
\caption{Numerical results of speech-to-text ST on MuST-C En$\rightarrow$Es with fixed pre-decision of 280$ms$.}
\label{enes_s2t}
\end{table*}

\begin{table*}[]
\centering
\small
\begin{tabular}{cccccc} \hlinew{1.2pt}
\multicolumn{6}{c}{\textbf{MuST-C English$\rightarrow$German}}              \\ \hline
\multicolumn{6}{c}{\textit{\textbf{Offline}}}            \\
        & CW      & AP   & AL      & DAL     & SacreBLEU \\
        & 5654.72 & 1.00 & 5654.72 & 5654.72 & 22.80     \\\hline
\multicolumn{6}{c}{\textit{\textbf{RealTranS}}}          \\
(K, N)  & CW      & AP   & AL      & DAL     & SacreBLEU \\
(3, 3)  & -       & -    & 1355    & -       & 16.54     \\
(5, 3)  & -       & -    & 1838    & -       & 18.49     \\
(7, 3)  & -       & -    & 2290    & -       & 19.84     \\
(9, 3)  & -       & -    & 2720    & -       & 20.05     \\
(11, 3) & -       & -    & 3106    & -       & 20.41     \\\hline
\multicolumn{6}{c}{\textit{\textbf{MoSST}}}              \\
$k$       & CW      & AP   & AL      & DAL     & SacreBLEU \\
1       & -       & 0.29 & 208     & 642     & 1.35      \\
3       & -       & 0.53 & 818     & 1182    & 6.75      \\
5       & -       & 0.79 & 1734    & 2263    & 16.34     \\
7       & -       & 0.93 & 2551    & 3827    & 19.77     \\
9       & -       & 0.96 & 2742    & 4278    & 19.97     \\\hline
\multicolumn{6}{c}{\textit{\textbf{ITST}}}               \\
$\delta$   & CW      & AP   & AL      & DAL     & SacreBLEU \\
0.75    & 558.30  & 0.73 & 1448.53 & 1720.45 & 17.90     \\
0.80    & 684.79  & 0.75 & 1588.52 & 2047.05 & 18.47     \\
0.81    & 773.64  & 0.77 & 1677.98 & 2251.77 & 19.09     \\
0.82    & 877.89  & 0.79 & 1778.44 & 2499.23 & 19.50     \\
0.83    & 1042.91 & 0.81 & 1918.86 & 2819.91 & 20.09     \\
0.84    & 1275.87 & 0.83 & 2136.53 & 3213.04 & 20.64     \\
0.85    & 1539.91 & 0.86 & 2370.87 & 3594.93 & 21.06     \\
0.86    & 1842.74 & 0.88 & 2617.66 & 3944.18 & 21.64     \\
0.87    & 2171.43 & 0.90 & 2892.93 & 4258.03 & 21.80     \\
0.88    & 2559.36 & 0.92 & 3192.52 & 4544.17 & 22.02     \\
0.89    & 2971.17 & 0.94 & 3501.27 & 4786.36 & 22.27     \\
0.90    & 3430.62 & 0.95 & 3875.92 & 5006.06 & 22.51     \\
0.92    & 4296.22 & 0.98 & 4556.58 & 5317.75 & 22.62     \\
0.95    & 5114.86 & 0.99 & 5206.45 & 5543.74 & 22.71    \\ \hlinew{1.2pt}
\end{tabular}
\caption{Numerical results of speech-to-text ST on MuST-C En$\rightarrow$De with flexible pre-decision on each speech frame. Note that `Offline' applies original Wav2Vec2.0 and unidirectional encoder, and ITST applies unidirectional Wav2Vec2.0 and unidirectional encoder.}
\label{ende_s2t_flexible}
\end{table*}

\end{document}